\documentclass[10pt]{article}
\newif\iffastcompile
\fastcompilefalse
\iffastcompile
  \PassOptionsToPackage{draft}{graphicx}
\fi
\usepackage{visualevolve-arxiv}
\usepackage{adjustbox}
\usepackage[round,authoryear]{natbib}
\usepackage{url}
\usepackage{booktabs}
\usepackage{multirow}
\usepackage{makecell}
\usepackage{array}
\usepackage{amsmath,amssymb}

\usepackage{graphicx}
\usepackage{capt-of}
\usepackage{fvextra}
\usepackage{needspace}

\usepackage{placeins}
\usepackage{enumitem}
\usepackage[ruled,vlined,linesnumbered]{algorithm2e}
\SetKwInOut{Require}{Require}
\SetKwInOut{Ensure}{Ensure}
\SetKw{Return}{return}
\providecommand{\Description}[1]{}

\DefineVerbatimEnvironment{PromptBlock}{Verbatim}{
  breaklines=true,
  breakanywhere=false,
  breakbefore={/},
  breakafter={,},
  breakindent=0.8em,
  breaksymbolleft={},
  breaksymbolright={},
  fontsize=\small,
  baselinestretch=0.88,
  frame=single,
  framerule=0.3pt,
  rulecolor=\color{black!35},
  fillcolor=\color{black!2},
  framesep=1.5mm
}

\newcounter{promptlisting}
\newcommand{\PromptListingTitle}[1]{%
  \Needspace{5\baselineskip}%
  \refstepcounter{promptlisting}%
  \par\addvspace{0.55\baselineskip}%
  \noindent\begin{minipage}{\linewidth}
    \centering\small\sffamily\bfseries\color{PaperInk}
    Listing~\thepromptlisting: #1
  \end{minipage}%
  \par\nobreak\vspace{2pt}%
}

\newcommand{\sys}{VisualEvolve}

\usepackage{wrapfig}
\usepackage{tocloft}
\usepackage[colorlinks=true,linkcolor=PaperAccent,citecolor=PaperAccent,urlcolor=PaperAccent,bookmarksnumbered=true]{hyperref}
\usepackage{bookmark}
\author{%
  Jiawei Wang\textsuperscript{1}, Yucun Li\textsuperscript{2}, Xueqi Li\textsuperscript{3}, Shenghao Wang\textsuperscript{2}, Xinquan Gong\textsuperscript{2},
  Dan Zhang\textsuperscript{1}, Zheqi Lv\textsuperscript{4}, Yuquan Le\textsuperscript{5}, Shaozu Yuan\textsuperscript{6}, An Zhang\textsuperscript{7}, Da Cao\textsuperscript{2,*}, Tat-Seng Chua\textsuperscript{1}
}
\renewcommand{\PaperAffiliations}{%
  \textsuperscript{1}National University of Singapore\quad
  \textsuperscript{2}Hunan University\quad
  \textsuperscript{3}Shenzhen University\\
  \textsuperscript{4}Cornell University\quad
  \textsuperscript{5}Hunan Normal University\quad
  \textsuperscript{6}Meituan\\
  \textsuperscript{7}University of Science and Technology of China\par
  {\small * Corresponding author.\par}
}
\renewcommand{\PaperAuthorMetadata}{Jiawei Wang, Yucun Li, Xueqi Li, Shenghao Wang, Xinquan Gong, Dan Zhang, Zheqi Lv, Yuquan Le, Shaozu Yuan, An Zhang, Da Cao, Tat-Seng Chua}

\hypersetup{pdftitle={VisualEvolve: A Skill Self-Evolution Framework for Visual Reasoning},pdfauthor={\PaperAuthorMetadata},pdfsubject={Visual reasoning through external skill self-evolution},pdfkeywords={Visual reasoning, vision-language models, skill evolution, training-free agents}}
\setlist[itemize]{leftmargin=1.3em,itemsep=3pt,topsep=4pt,parsep=0pt}
\begin{document}
\raggedbottom
\title{VisualEvolve: A Skill Self-Evolution Framework for Visual Reasoning}

\addtocontents{toc}{\protect\setcounter{tocdepth}{-1}}

\begin{abstract}
Visual tool use enables vision-language models (VLMs) to acquire task-relevant evidence, but its effectiveness can be limited by predefined visual operations and the tool-use strategies learned during training. Representing these operations and strategies as reusable external skills allows them to be refined through self-evolution without updating model parameters. However, skill refinement can introduce performance regressions. Meanwhile, inspecting accumulated task execution trajectories together with skill evolution history may lead to evolution evidence overload. To address these challenges, we propose VisualEvolve, a self-evolving framework that evolves executable visual operations and tool-use strategies for VLMs through a Select--Execute--Evolve--Evaluate cycle. Skill-Library Snapshot Selection uses performance on a fixed validation set disjoint from execution tasks to select an effective historical snapshot for subsequent evolution, limiting the propagation of performance regressions. Progressive Evidence Access (PEA) guides capability-gap diagnosis through a coarse-to-fine hierarchy comprising the Skill-Performance Report, Evolution Log, Trajectory Summaries, and relevant raw trajectory segments. Across four VLM backbones and six evaluation sets from three benchmarks, VisualEvolve improves over the Initial Skill Library in 23 of 24 settings, with a maximum gain of 15.1 percentage points on VisualProbe Hard using Qwen3.5-Plus. Ablation studies support the contributions of Snapshot Selection and PEA, while further experiments show that skills can be improved using experience from different data settings and directly reused across backbones. Moreover, VisualEvolve enables effective skill self-evolution with over an order of magnitude fewer task-solving trajectories than estimated for representative RL-based methods. 

\end{abstract}

\maketitle
\begin{figure}[!t]
  \centering
  \includegraphics[width=\linewidth]{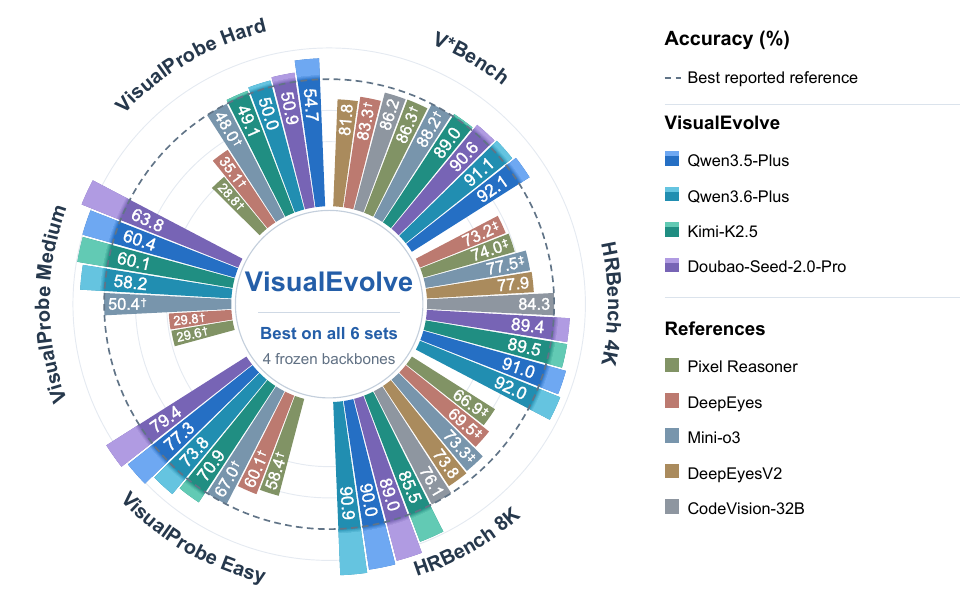}
  \caption{\textbf{VisualEvolve versus training-based agents on six evaluation sets.} VisualEvolve uses our framework with four backbones; References are agents trained with SFT and/or RL. The dashed circle marks the highest reported reference accuracy in each set. VisualEvolve: Acc@1; $\dagger$: Avg@32; $\ddagger$: Avg@8.}
  \label{fig:teaser}
\end{figure}

\section{Introduction}
\label{sec:intro}

By adopting visual tool use, vision-language models (VLMs) are capable of inspecting and manipulating images during reasoning~\citep{vstar,vtoolr1,codedance}.
Its effectiveness depends on both the available visual operations and the strategies that determine when and how to use them~\citep{deepeyes,minio3}.
Training-based approaches, such as DeepEyes~\citep{deepeyes} and Mini-o3~\citep{minio3}, learn these strategies through supervised fine-tuning (SFT) or reinforcement learning (RL) over predefined visual operations.
However, limitations in available visual operations or their usage strategies may prevent models from acquiring task-relevant visual evidence~\citep{minio3,dynamo,monet}.
This motivates adapting both visual operations and tool-use strategies to the needs revealed during task execution~\citep{dynamo,meta,adaptvision}.
Recent work represents reusable task guidance as external skill -- structured instructions for performing a class of tasks, which can be paired with executable tools~\citep{dynamo,polyskill}.
Stored outside model parameters, these instructions and associated tools can be refined without retraining the model~\citep{dynamo}.

Self-evolution enables an agent to refine its external skill library through task execution and feedback~\citep{xskill,dynamo}.
The goal is to turn experience accumulated over multiple rounds into reusable visual reasoning skills~\citep{xskill,trace2skill,dynamo}.
However, skill evolution faces two challenges.
First, skill refinement can introduce performance regressions, even when it addresses failures observed during task execution~\citep{skillsmith,skillhone}.
Second, skill evolution requires analyzing not only task execution trajectories, but also the history of skill modifications and evaluation results~\mbox{\citep{skillhone,skillsmith}}.
As these records accumulate, inspecting them all at once can cause evolution evidence overload, increasing the context burden and making capability gaps harder to diagnose~\citep{metaharness,museautoskill}.

To address these challenges, we propose VisualEvolve, a self-evolving framework that evolves executable visual operations and tool-use strategies for VLMs through a Select--Execute--Evolve--Evaluate cycle. Each cycle comprises four stages: (1) Select. To limit the propagation of performance regressions, Skill-Library Snapshot Selection guides the agent to choose a starting library from archived snapshots based on validation performance; (2) Execute. The selected library is applied to visual reasoning tasks, producing execution trajectories, correctness labels, and Trajectory Summaries for capability-gap diagnosis; (3) Evolve. To address evolution evidence overload, we introduce Progressive Evidence Access (PEA), which guides capability-gap diagnosis through coarse-to-fine access to skill evolution history and current execution evidence. These diagnoses guide revisions to the skill library; (4) Evaluate. The revised library is assessed on a fixed validation set disjoint from the execution tasks and archived as a new snapshot. The resulting validation feedback and updated evolution records support subsequent snapshot selection and capability-gap diagnosis.
Together, these components enable VisualEvolve to build on effective historical skill libraries while diagnosing capability gaps through coarse-to-fine access to execution evidence and evolution history. 

As summarized in Figure~\ref{fig:teaser}, experiments with four frozen VLM backbones across six evaluation sets from three benchmarks show that evolving external skills improves visual reasoning without updating model parameters.
VisualEvolve outperforms the Initial Skill Library in 23 of 24 backbone--evaluation-set combinations, with the largest gain of 15.1 percentage points (38.1\% relative) on VisualProbe Hard using Qwen3.5-Plus.
Ablation studies show that both selecting the skill-library snapshot for further evolution and organizing evidence through PEA contribute to these gains.
Experiments under three evolution-data settings further show that VisualEvolve can improve its skills through self-evolution using execution experience from different data sources.
Finally, we transfer the skill library evolved with Kimi-K2.5 to three other VLM backbones without modification. The transferred library improves their average accuracy over the Initial Skill Library, demonstrating that the evolved skills can be directly reused across models. Our main contributions can be summarized as follows:

\begin{itemize}
    \item We propose VisualEvolve, a self-evolving framework that evolves executable visual operations and tool-use strategies for VLMs through a Select--Execute--Evolve--Evaluate cycle.

    \item We introduce Skill-Library Snapshot Selection to select the historical snapshot with the highest validation accuracy for subsequent evolution, and Progressive Evidence Access to guide capability-gap diagnosis through coarse-to-fine access to task execution trajectories and skill evolution history.

    \item We demonstrate that VisualEvolve improves visual reasoning across four backbones and three benchmarks without updating model parameters. Ablation studies demonstrate the contributions of key stages in the evolution cycle and Progressive Evidence Access. Further experiments demonstrate effective skill improvement through self-evolution using experience from different data sources and direct reuse of evolved skills across backbones.
\end{itemize}

\section{Related Work}
\label{sec:related}

\noindent\textbf{Training-Based Visual Reasoning Agents.}
Training-based visual reasoning agents acquire task-relevant evidence
through tool use, from invoking predefined tools such as
cropping~\citep{minio3,lfpc,sensesearch} to generating and executing
image-processing code~\citep{thyme,codedance,codevision}. These methods typically optimize when and how to invoke such operations through SFT, RL, or both~\citep{pixelreasoner,vistar1,spacetools}. In contrast, \sys{} keeps the backbone frozen and improves visual reasoning by evolving external skills.

\noindent\textbf{Skill-Based Agents.}
External skills encode reusable capabilities~\citep{polyskill} as executable programs~\citep{voyager,sage,nsi}, structured guidance for planning and tool use~\citep{xskill}, or natural-language instructions augmented with optional executable and supporting resources~\citep{trace2skill,museautoskill}. \sys{} adapts this paradigm to visual reasoning by representing each skill as a tool-use strategy paired with an optional visual operation.

\noindent\textbf{Self-Evolving Agents.}
Self-evolving agents improve over time by externalizing accumulated experience into reusable artifacts, including verbal reflections~\citep{reflexion}, experiential knowledge~\citep{expel,reasoningbank}, and reusable skills~\citep{voyager,skillsmith,trace2skill,dynamo}. Recent methods retain richer evolution histories. SkillHone records diagnoses, revisions, and feedback to support continual skill refinement~\citep{skillhone}, while Meta-Harness gives its proposer access to the source code, scores, and traces of prior harness candidates~\citep{metaharness}. \sys{} focuses on two challenges in multi-round skill evolution: performance regressions after skill updates and evidence overload from accumulated evolution history. Snapshot Selection selects the historical snapshot with the highest validation accuracy for subsequent evolution, while Progressive Evidence Access organizes accumulated evidence into a coarse-to-fine hierarchy for capability-gap diagnosis.

\begin{figure*}[!t]
  \centering
  \includegraphics[width=\textwidth]{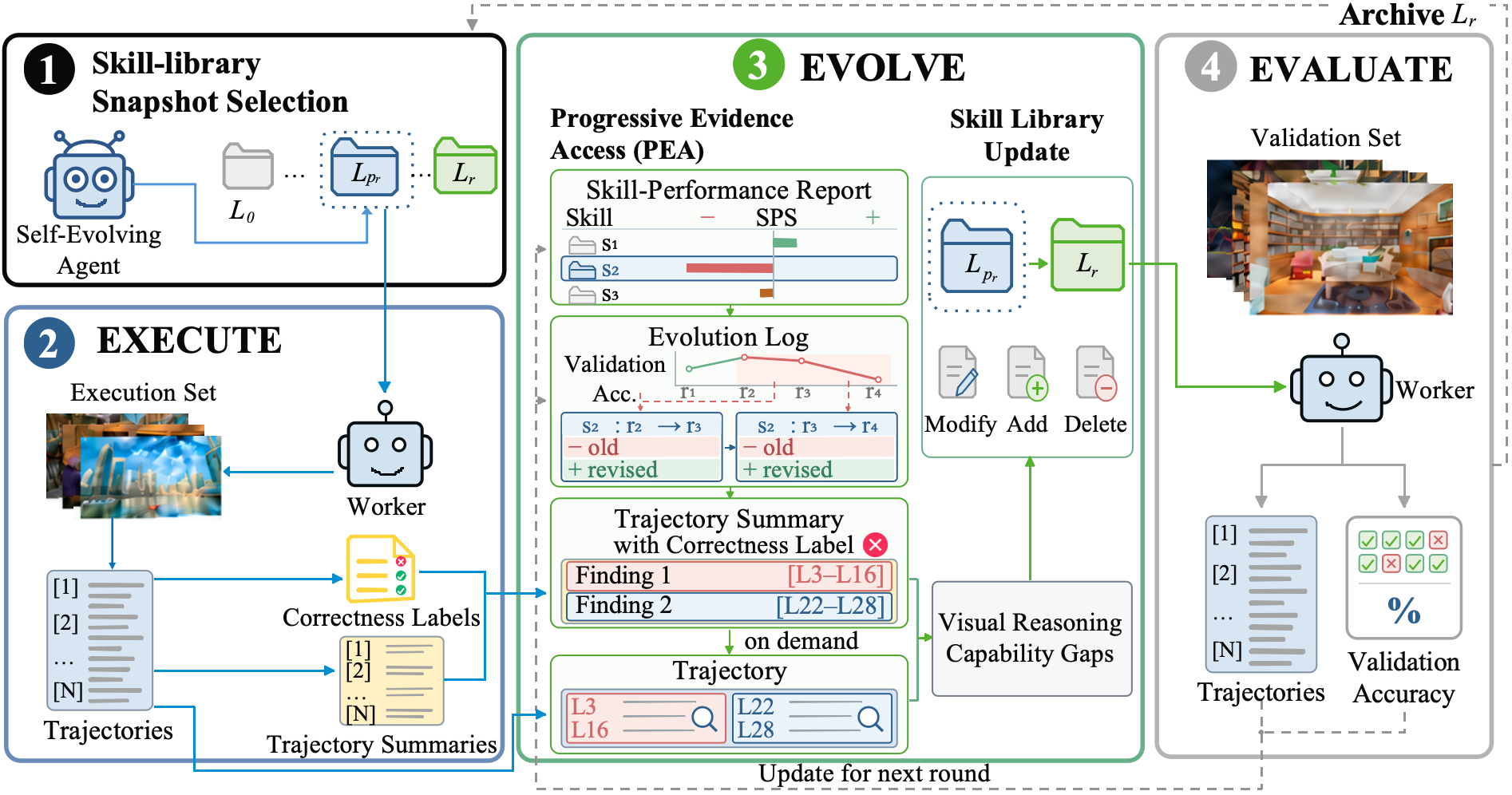}
  \caption{\textbf{The \sys{} framework.} \sys{} evolves a skill library through multi-round Select--Execute--Evolve--Evaluate cycles.}
  \Description{The Self-Evolving Agent follows stage-specific evolution
  Meta-skill prompts to execute tasks, construct evidence-linked Memories,
  update an external skill library, and evaluate each updated snapshot.}
  \label{fig:framework}
\end{figure*}

\section{Method}
\label{sec:method}

\subsection{Problem Formulation}
\label{sec:method-problem}
We study the problem of evolving a skill library to improve performance
on visual reasoning tasks. Given a frozen backbone
$\mathcal{A}_{\theta}$, an initial skill library $\mathcal{L}_0$, evolution
data $\mathcal{D}_{\mathrm{evo}}$, and an evolution budget of $R$ rounds, we
construct a fixed validation set
$\mathcal{D}_{\mathrm{val}}\subset\mathcal{D}_{\mathrm{evo}}$ and
round-specific execution sets
$\mathcal{D}^{\mathrm{exe}}_r\subset
\mathcal{D}_{\mathrm{evo}}\setminus\mathcal{D}_{\mathrm{val}}$ for
$r=1,\ldots,R$. \sys{} iteratively updates the skill library and obtains a
sequence of skill-library snapshots
$\{\mathcal{L}_0,\mathcal{L}_1,\ldots,\mathcal{L}_R\}$.
Each snapshot $\mathcal{L}_t$ is evaluated on
$\mathcal{D}_{\mathrm{val}}$, yielding the validation accuracy
$a_t^{\mathrm{val}}=\mathrm{Acc}(\mathcal{A}_{\theta},\mathcal{L}_t;
\mathcal{D}_{\mathrm{val}})$. We select the snapshot with the highest validation accuracy as the final
skill library $\mathcal{L}^{\star}$ and assess its performance on
evaluation sets reserved for final testing.

\subsection{Framework Overview}
\label{sec:method-overview}

As shown in Figure~\ref{fig:framework}, \sys{} evolves a skill library through multi-round Select--Execute--Evolve--Evaluate cycles. Skill-Library Snapshot Selection chooses the starting skill library, Execute applies it to the execution set to collect execution evidence, Evolve diagnoses capability gaps and updates the skill library, and Evaluate assesses and archives the updated skill library. The prompt guiding the evolution process is provided in Appendix~\ref{app:self-evolving-agent-prompt}.

\subsection{Skill-Library Snapshot Selection} 
\label{sec:method-branchable-evolution} 

To limit the propagation of performance regressions across evolution rounds, each round begins with Skill-Library Snapshot Selection, which selects a previously archived skill-library snapshot as the starting point. Formally, we denote a skill library by $\mathcal{L}=\{s_i\}_{i=1}^{N}$, where each skill  $s_i=(\pi_i,u_i)$ consists of a textual strategy $\pi_i$ and an optional executable visual operation $u_i$. 

In the first round, \sys{} starts from the initial skill library $\mathcal{L}_0$, which contains a single crop skill and is detailed in Appendix~\ref{app:initial-skill-library}. For each subsequent round $r>1$, rather than always evolving from the latest snapshot, \sys{} selects the archived snapshot with the highest validation accuracy among $\mathcal{L}_0,\ldots,\mathcal{L}_{r-1}$. The selected snapshot, denoted by $\mathcal{L}_{p_r}$, serves as the starting skill library for round $r$.

\subsection{Execute}
\label{sec:method-execute}

To construct execution evidence for Evolve, Execute applies the selected starting skill library $\mathcal{L}_{p_r}$ to the execution set, producing trajectories, correctness labels, and Trajectory Summaries.

At round $r$, the Worker, a task-solving agent instantiated with the frozen backbone $\mathcal{A}_{\theta}$ and equipped with $\mathcal{L}_{p_r}$, solves tasks from the execution set $\mathcal{D}_r^{\mathrm{exe}}$, producing trajectories that record the task-solving process and final predictions. To distinguish successful and failed outcomes for subsequent diagnosis, each trajectory is assigned a correctness label by comparing its final prediction with the ground-truth answer. Finally, to reduce the context burden of trajectory analysis while preserving traceable evidence, each trajectory is compressed into a source-linked Trajectory Summary, where each summary comprises multiple findings that capture capability-relevant behaviors and reference the corresponding supporting spans in the raw trajectory. Additional details and the Execute instruction are provided in Appendices~\ref{app:execute-details} and~\ref{app:execute-meta-skill}, respectively.

\subsection{Evolve} \label{sec:method-evolve} 
Evolve uses execution evidence together with the history of skill modifications and evaluation results to diagnose capability gaps and update the starting skill library $\mathcal{L}_{p_r}$. Since loading all available evidence at once would incur evolution evidence overload, we introduce Progressive Evidence Access, which enables the Self-Evolving Agent, primarily responsible for capability-gap diagnosis and skill-library updates, to access evidence from coarse to fine. It first identifies relevant skills through the Skill-Performance Report, traces their historical modifications through the Evolution Log, analyzes current-round Trajectory Summaries for recurring failure patterns, and accesses relevant raw trajectory segments only when finer-grained evidence is needed.

\noindent\textbf{Skill-Performance Report.}
Validation accuracy measures the performance of a skill library but does not indicate which skills are associated with performance changes. We therefore introduce the Skill-Performance Report, which estimates each skill's association with changes in validation performance through a Skill Performance Score (SPS). The SPS is defined as the difference in mean correctness between two skill-library snapshots over the skill-associated validation subset.

Specifically, for two snapshots $\mathcal{L}_a$ and
$\mathcal{L}_b$, we define the skill-associated validation subset as
\[
\mathcal{D}^{a,b}_s =
\left\{
d_j\in\mathcal{D}_{\mathrm{val}}
\mid
s\in Z_j^{(a)} \lor s\in Z_j^{(b)}
\right\},
\]
where $Z_j^{(a)}$ and $Z_j^{(b)}$ denote the sets of skills loaded while
solving validation example $d_j$ with $\mathcal{L}_a$ and
$\mathcal{L}_b$, respectively. Thus, $\mathcal{D}^{a,b}_s$ contains all
validation examples on which $s$ is loaded under at least one of the two
snapshots. For $|\mathcal{D}^{a,b}_s|>0$, SPS is computed as the difference in mean
correctness between the two snapshots on this subset:
\[
\operatorname{SPS}_{a\rightarrow b}(s)
=
\frac{1}{|\mathcal{D}^{a,b}_s|}
\sum_{d_j\in\mathcal{D}^{a,b}_s} c^{(b)}(d_j)
-
\frac{1}{|\mathcal{D}^{a,b}_s|}
\sum_{d_j\in\mathcal{D}^{a,b}_s} c^{(a)}(d_j),
\]
where $c^{(a)}(d_j),c^{(b)}(d_j)\in\{0,1\}$ indicate whether $d_j$ is
answered correctly under $\mathcal{L}_a$ and $\mathcal{L}_b$, respectively.
The sign of SPS indicates the direction of the performance difference, while its absolute value indicates the strength of association. Since multiple skills may be loaded together, SPS provides only an approximate estimate of an individual skill's contribution to the performance change.

At round $r>1$, the Self-Evolving Agent uses the report from round $r-1$,
which contains SPS values for up to three snapshot pairs:
$(\mathcal{L}_{p_{r-1}},\mathcal{L}_{r-1})$,
$(\mathcal{L}_0,\mathcal{L}_{r-1})$, and
$(\mathcal{L}_0,\mathcal{L}_{p_{r-1}})$.
The agent considers these values jointly and prioritizes skills with large
absolute SPS for further inspection.

\noindent\textbf{Evolution Log.}
The Skill-Performance Report identifies skills associated with performance changes but does not reveal how these skills have changed across archived snapshots. We therefore introduce the Evolution Log, which records each snapshot's validation accuracy and indexes per-skill differences between consecutive snapshots. The Self-Evolving Agent uses the accuracy history to locate improvement or regression intervals and inspects the corresponding historical changes of the identified skills, providing references for update directions to reuse or avoid.

\noindent\textbf{Trajectory Summary.}
After obtaining historical references for skill updates, the Self-Evolving Agent turns to current-round execution evidence. Using the correctness labels, it prioritizes the Trajectory Summaries associated with incorrect predictions, identifies recurring failure patterns, and abstracts them into capability gaps, such as repeatedly revisiting the same image regions and stopping visual search before sufficient evidence is collected.

\noindent\textbf{Trajectory.}
When finer-grained evidence is needed, the Self-Evolving Agent follows the
source references in Trajectory Summaries to relevant segments of
trajectories, exposing detailed reasoning and tool-use behavior behind the
identified failures.

Based on the diagnosed capability gaps, Evolve updates the starting skill library $\mathcal{L}_{p_r}$ by modifying, adding, or removing skills, yielding $\mathcal{L}_r$. For an existing skill $s_i=(\pi_i,u_i)$, the update can revise its tool-use strategy $\pi_i$ or executable visual operation $u_i$. Before Evaluate, $\mathcal{L}_r$ undergoes structural and executability checks. These checks and the construction of the Skill-Performance Report and Evolution Log are detailed in Appendix~\ref{app:evolve-details}, while the Evolve instruction is provided in Appendix~\ref{app:evolve-meta-skill}.

\subsection{Evaluate}
\label{sec:method-evaluate}

To provide a consistent generalization signal across evolution rounds, Evaluate assesses the updated skill library $\mathcal{L}_r$ on the fixed validation set $\mathcal{D}_{\mathrm{val}}$ using the Worker, yielding the validation accuracy $a_r^{\mathrm{val}}$. The Skill-Performance Report and Evolution Log are then updated for use in the next round's Evolve stage, while $\mathcal{L}_r$ is archived as a new skill-library snapshot for the next round's Snapshot Selection. The instruction for Evaluate is provided in Appendix~\ref{app:evaluate-meta-skill}.

\subsection{Evolution Procedure}
\label{sec:method-procedure}
The complete evolution procedure repeatedly cycles through Skill-Library Snapshot Selection, Execute, Evolve, and Evaluate until all $R$ evolution rounds are completed. All agents use the same frozen VLM backbone. At each round $r$, Snapshot Selection first determines the starting skill library $\mathcal{L}_{p_r}$. Execute applies $\mathcal{L}_{p_r}$ to a round-specific
execution set, producing trajectories, correctness labels, and Trajectory Summaries. Evolve then uses Progressive Evidence Access to diagnose capability gaps using the Skill-Performance Report, Evolution Log, Trajectory Summaries, and, when needed, relevant raw trajectory segments, and updates the starting skill library to produce the updated library $\mathcal{L}_r$. Evaluate assesses and archives the $\mathcal{L}_r$, while updating the Skill-Performance Report and Evolution Log for subsequent rounds.
After all $R$ rounds are completed, the archived snapshot with the highest
validation accuracy is returned as the final skill library
$\mathcal{L}^{\star}$. Algorithm~\ref{alg:visualevolve} summarizes the complete evolution procedure.

\section{Experiments}
\label{sec:exp}

To evaluate VisualEvolve, we conduct comprehensive experiments to explore the following research questions. \textbf{RQ1:} How does VisualEvolve improve over Single-Pass inference and the Initial Skill Library, and how does it compare with representative training-based visual-reasoning baselines? \textbf{RQ2:} How do the framework stages and the evidence-access mechanism within Progressive Evidence Access contribute to effective multi-round skill evolution? \textbf{RQ3:} How does VisualEvolve evolve a skill library using execution experience from different data sources, and how well does the evolved skill library transfer across backbones? \textbf{RQ4:} What is the cost of skill evolution, and what additional inference-time cost does the evolved skill library introduce?

\subsection{Experimental Setup} \label{sec:exp-setup}

\noindent\textbf{Evolution Data.}
We use evolution data $\mathcal{D}_{\mathrm{evo}}$ as the source of execution tasks for skill evolution. In the default setting, $\mathcal{D}_{\mathrm{evo}}$ combines VisualProbe$_{\mathrm{train}}$, with 2.9K samples for challenging visual search~\citep{minio3}, and the Visual Search component of the DeepEyes training corpus, with approximately 22K samples from the $V^*$ training set for visual grounding and fine-grained perception~\citep{vstar,deepeyes}. We retain only the original image, question, and answer.

\noindent\textbf{Benchmarks.}
We use separate benchmarks for final evaluation. V\textsuperscript{*}Bench contains 191 examples for fine-grained visual search and spatial reasoning~\citep{vstar}, HRBench4K and HRBench8K each contain 800 high-resolution visual-reasoning examples~\citep{hrbench}, and VisualProbe Easy, Medium, and Hard contain 141, 268, and 106 examples, respectively, covering increasingly challenging exploratory visual reasoning~\citep{minio3}.

\noindent\textbf{Baselines.}
We compare \sys{} against three types of baselines. Single-Pass queries the VLM backbone in a single turn without external tools, serving as the baseline. Initial Skill Library equips each backbone with $\mathcal{L}_0$ and serves as the baseline for measuring gains from evolution. Training-Based Methods include representative visual-reasoning agents trained with RL, with or without prior SFT~\citep{pixelreasoner,deepeyes,minio3,deepeyesv2,codevision}.

\noindent\textbf{Implementation Details.}
We run \sys{} for $R=16$ evolution rounds. Before the first round, we sample a fixed 200-example validation set $\mathcal{D}_{\mathrm{val}}$ from $\mathcal{D}_{\mathrm{evo}}$. At each round $r$, we sample 32 examples from $\mathcal{D}_{\mathrm{evo}}\setminus\mathcal{D}_{\mathrm{val}}$ to form $\mathcal{D}_r^{\mathrm{exe}}$. When $\mathcal{D}_{\mathrm{evo}}$ combines multiple datasets, balanced sampling is applied to both the validation and execution sets, drawing an equal number of examples from each dataset. We use Qwen3.5-Plus~\citep{qwen35}, Qwen3.6-Plus~\citep{qwen36}, Kimi-K2.5~\citep{kimik25}, and Doubao-Seed-2.0-Pro~\citep{seed20}, with temperature $0$ and thinking mode enabled.

\begin{table}[!t]
\centering
\normalsize
\setlength{\belowcaptionskip}{\baselineskip}

\fontsize{9}{11}\selectfont
\setlength{\tabcolsep}{2.2pt}
\setlength{\aboverulesep}{1.6pt}
\setlength{\belowrulesep}{1.8pt}
\renewcommand{\arraystretch}{1.0}
\begin{tabular*}{\linewidth}{@{\extracolsep{\fill}}llrrrrrrr@{}}
\toprule
\multirow{2}{*}{\textbf{Backbone}} & \multirow{2}{*}{\textbf{Method}}
& \multirow{2}{*}{\makecell{\textbf{V\textsuperscript{*}}\\\textbf{Bench}}}
& \multicolumn{2}{c}{\textbf{HRBench}}
& \multicolumn{3}{c}{\textbf{VisualProbe}}
& \multirow{2}{*}{\makecell{\textbf{Macro}\\\textbf{Avg.}}} \\
\cmidrule(lr){4-5}\cmidrule(lr){6-8}
& & & \textbf{4K} & \textbf{8K}
& \textbf{Easy} & \textbf{Medium} & \textbf{Hard} & \\
\midrule
\multirow{4}{*}{Qwen2.5-VL-7B}
& Pixel Reasoner$^{\mathrm{R}}$ & \underline{86.3}$^\dagger$ & 74.0$^\ddagger$ & 66.9$^\ddagger$ & 58.4$^\dagger$ & 29.6$^\dagger$ & 28.8$^\dagger$ & 57.3 \\
& DeepEyes$^{\mathrm{R}}$ & 83.3$^\dagger$ & 73.2$^\ddagger$ & 69.5$^\ddagger$ & \underline{60.1}$^\dagger$ & \underline{29.8}$^\dagger$ & \underline{35.1}$^\dagger$ & \underline{58.5} \\
& Mini-o3 & \textbf{88.2}$^\dagger$ & 77.5$^\ddagger$ & 73.3$^\ddagger$ & \textbf{67.0}$^\dagger$ & \textbf{50.4}$^\dagger$ & \textbf{48.0}$^\dagger$ & \textbf{67.4} \\
& DeepEyesV2 & 81.8 & \underline{77.9} & \underline{73.8} & -- & -- & -- & -- \\
Qwen3-VL-32B & CodeVision-32B & 86.2 & \textbf{84.3} & \textbf{76.1} & -- & -- & -- & -- \\
\midrule
\multirow{3}{*}{Qwen3.5-Plus}
& Single-Pass & 83.8 & 84.4 & 80.0 & 66.7 & 37.7 & 35.8 & 64.7 \\
& Initial Skill Library & \underline{86.4} & \underline{87.0} & \underline{83.3} & \underline{68.1} & \underline{48.1} & \underline{39.6} & \underline{68.8} \\
& \textbf{\sys{}} & \textbf{92.1} & \textbf{91.0}
& \textbf{90.0} & \textbf{77.3} & \textbf{60.4}
& \textbf{54.7} & \textbf{77.6} \\
\midrule
\multirow{3}{*}{Qwen3.6-Plus}
& Single-Pass & 88.0 & 87.8 & 84.1 & 58.9 & 34.7 & 31.1 & 64.1 \\
& Initial Skill Library & \underline{90.6} & \underline{88.5} & \underline{87.4} & \underline{66.7} & \underline{53.4} & \underline{36.8} & \underline{70.6} \\
& \textbf{\sys{}} & \textbf{91.1} & \textbf{92.0}
& \textbf{90.9} & \textbf{73.8} & \textbf{58.2}
& \textbf{50.0} & \textbf{76.0} \\
\midrule
\multirow{3}{*}{Kimi-K2.5}
& Single-Pass & 82.7 & 79.8 & 75.8 & \underline{50.4} & 32.1 & 24.5 & 57.6 \\
& Initial Skill Library & \underline{85.3} & \underline{88.5} & \underline{81.5} & \textbf{70.9} & \underline{54.1} & \underline{35.8} & \underline{69.4} \\
& \textbf{\sys{}} & \textbf{89.0} & \textbf{89.5}
& \textbf{85.5} & \textbf{70.9} & \textbf{60.1}
& \textbf{49.1} & \textbf{74.0} \\
\midrule
\multirow{3}{*}{\makecell[l]{Doubao-Seed-\\2.0-Pro}}
& Single-Pass & 88.5 & 83.3 & 76.5 & 60.3 & 36.2 & 29.2 & 62.3 \\
& Initial Skill Library & \underline{89.5} & \underline{86.8} & \underline{86.0} & \underline{75.9} & \underline{59.0} & \underline{42.5} & \underline{73.3} \\
& \textbf{\sys{}} & \textbf{90.6} & \textbf{89.4}
& \textbf{89.0} & \textbf{79.4} & \textbf{63.8}
& \textbf{50.9} & \textbf{77.2} \\
\bottomrule
\end{tabular*}
\par
\caption{\textbf{Main results (RQ1).} Accuracy (\%) on six evaluation sets from the three benchmarks. For training-based methods, results in rows marked with $^{\mathrm{R}}$
are taken from the Mini-o3 paper~\citep{minio3}, while results for the
remaining methods are taken from their respective original papers.}
\label{tab:benchmark}\par
\vspace{2pt}
\noindent\begin{minipage}{\linewidth}
\footnotesize\raggedright
Sources: Mini-o3~\citep{minio3}, DeepEyesV2~\citep{deepeyesv2}, and CodeVision~\citep{codevision}.\par

$\dagger$: Avg@32; $\ddagger$: Avg@8; our runs: Acc@1.
Macro Avg.: unweighted mean of all six sets, shown only for complete rows.
``--'': unavailable.
\textbf{Bold}/\underline{underline}: best/second-best per column within training-based methods or each of our backbone groups; ties share a rank.
\end{minipage}
\end{table}

\subsection{Main Results (RQ1)} \label{sec:exp-benchmark} 
We evaluate \sys{} with two internal baselines, Single-Pass and the Initial Skill Library, across four backbones, and further compare recent training-based visual-reasoning agents. We report accuracy on six evaluation sets from three benchmarks in Table~\ref{tab:benchmark}.

We make the following observations from Table~\ref{tab:benchmark}. 
(1) Multi-round skill evolution consistently improves performance over both Single-Pass and the Initial Skill Library. Across the 24 backbone--evaluation-set combinations, \sys{} outperforms Single-Pass in all cases and improves over the Initial Skill Library in 23 cases. The largest gains over the Initial Skill Library occur on VisualProbe Hard, reaching 15.1 and 13.3 percentage points with Qwen3.5-Plus and Kimi-K2.5, respectively. These results demonstrate the effectiveness of multi-round skill evolution.
(2) Skill evolution yields consistent gains across all four backbones. Compared with the Initial Skill Library, \sys{} improves macro-average accuracy by 8.8, 5.4, 4.6, and 3.9 percentage points with Qwen3.5-Plus, Qwen3.6-Plus, Kimi-K2.5, and Doubao-Seed-2.0-Pro, respectively. These consistent gains support the robustness of skill evolution across different backbones.
(3) We also include reported results from representative training-based visual-reasoning agents for reference. With Qwen3.5-Plus, \sys{} reaches 77.6\% macro-average accuracy, compared with 67.4\% for Mini-o3, the strongest training-based method with complete reported results, suggesting that \sys{} achieves competitive performance among the reported results.

\begin{figure*}[!t]
\centering

\begin{minipage}[c]{0.64\textwidth}
\centering

\vspace{1pt}

\begingroup
\fontsize{9pt}{9.5pt}\selectfont
\setlength{\tabcolsep}{1.6pt}
\setlength{\aboverulesep}{1pt}
\setlength{\belowrulesep}{1pt}
\renewcommand{\arraystretch}{0.90}
\newcommand{\kept}{$\checkmark$}
\newcommand{\removed}{$\times$}
\begin{tabular*}{\linewidth}{@{\extracolsep{\fill}}lccccccr@{}}
\toprule
& \multicolumn{3}{c}{\textbf{Stage}}
& \multicolumn{3}{c}{\textbf{PEA}} & \\
\cmidrule(lr){2-4}\cmidrule(lr){5-7}
\textbf{Variant} & Select
& Execute & Evaluate & SPR & EL & TS
& \makecell{\textbf{Macro}\\\textbf{Avg.}} \\
\midrule
\multicolumn{8}{@{}l}{\textit{Stage ablations}} \\
w/o Selection
& \removed & \kept & \kept & \kept & \kept & \kept & 72.2 \\
w/o Execute   & \kept & \removed & \kept & \kept & \kept & \removed & 69.9 \\
w/o Evaluate  & \removed & \kept & \removed & \removed & \removed & \kept & 71.2 \\
\midrule
\multicolumn{8}{@{}l}{\textit{PEA ablations}} \\
w/o SPR       & \kept & \kept & \kept & \removed & \kept & \kept & 73.6 \\
w/o EL        & \kept & \kept & \kept & \kept & \removed & \kept & 75.0 \\
w/o TS        & \kept & \kept & \kept & \kept & \kept & \removed & 73.0 \\
w/o SPR + EL  & \kept & \kept & \kept & \removed & \removed & \kept & 71.8 \\
w/o PEA       & \kept & \kept & \kept & \removed & \removed & \removed & 71.3 \\
\midrule
Single-Pass & \removed & \removed & \removed & \removed & \removed & \removed & 64.7 \\
\textbf{\sys{}} & \kept & \kept & \kept & \kept & \kept & \kept & \textbf{77.6} \\
\bottomrule
\end{tabular*}
\par
\captionof{table}{\textbf{Ablation of framework stages and Progressive Evidence Access (RQ2).}
All variants use Qwen3.5-Plus. Macro Avg. is the mean Acc@1 (\%) across six evaluation sets.}
\label{tab:pma-ablation-benchmarks}

\vspace{2pt}
\begin{minipage}{\linewidth}
\fontsize{9pt}{10pt}\selectfont
\raggedright
\textit{Notation:} \kept\ retained; \removed\ removed.
Select: Skill-Library Snapshot Selection;
SPR: Skill-Performance Report; EL: Evolution Log; TS: Trajectory Summary.
For w/o Evaluate, we evaluate all snapshots on the fixed
validation set only after evolution, then test the highest-scoring
snapshot on held-out benchmarks.
\end{minipage}
\endgroup
\end{minipage}
\hfill
\begin{minipage}[c]{0.32\textwidth}
\centering
\includegraphics[width=\linewidth,trim=1bp 10bp 14bp 6bp,clip]
{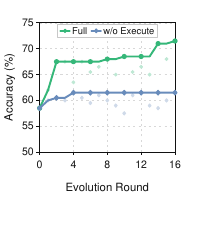}
\setlength{\abovecaptionskip}{4pt}
\raggedright
\captionsetup{justification=raggedright}
\captionof{figure}{\textbf{Execute ablation (RQ2).} Lines track the best-so-far accuracy over 16 rounds. Scatter points show each round's validation accuracy.}
\Description{A plot compares the full method with the variant that removes Execute
 over 16 evolution rounds.}
\label{fig:execute-evaluate-rq2}
\end{minipage}

\end{figure*}

\subsection{Ablation Study (RQ2)}
\label{sec:exp-ablation}

We conduct ablation studies with Qwen3.5-Plus to assess framework stages and the evidence-access mechanism in Progressive Evidence Access.

For the stage ablations, \textit{w/o Skill-Library Snapshot Selection} starts from the latest skill library. \textit{w/o Execute} removes execution evidence. \textit{w/o Evaluate} removes evaluation on the validation set, thereby disabling Snapshot Selection and leaving the Skill-Performance Report and Evolution Log empty. For Progressive Evidence Access ablations, \textit{w/o Skill-Performance Report} removes skill screening before Evolution Log analysis, \textit{w/o Evolution Log} removes the structured association between historical skill modifications and validation accuracy, and \textit{w/o Trajectory Summary} removes trajectory compression. \textit{w/o Skill-Performance Report + Evolution Log} removes the hierarchical access provided by these two components, while \textit{w/o Progressive Evidence Access} further removes Trajectory Summaries, so validation-accuracy history, skill-change history, and raw trajectories are accessed directly without hierarchical organization.

We make the following observations from Table~\ref{tab:pma-ablation-benchmarks} and Figure~\ref{fig:execute-evaluate-rq2}. (1) All three framework stages contribute substantially to
multi-round skill evolution. Compared with the full framework, removing Skill-Library Snapshot Selection, Execute, or Evaluate reduces macro-average accuracy by
5.4, 7.7, and 6.4 percentage points, respectively. In particular, \textit{w/o Execute}
reaches only 61.5\% best-so-far validation accuracy, highlighting the
importance of execution evidence for effective skill updates.
(2) The ablations support the effectiveness of the coarse-to-fine evidence-access mechanism in Progressive Evidence Access. Removing any individual component reduces macro-average accuracy by 2.6--4.6 percentage points. The combined ablations lead to larger degradation: jointly removing the Skill-Performance Report and Evolution Log reduces accuracy by 5.8 percentage points, while removing PEA entirely leads to the largest drop of 6.3 percentage points.

\subsection{Generalization (RQ3)}
\label{sec:exp-generalization}

\begin{figure}[!t]
\centering
\captionsetup{skip=4pt}
\includegraphics[width=\linewidth]{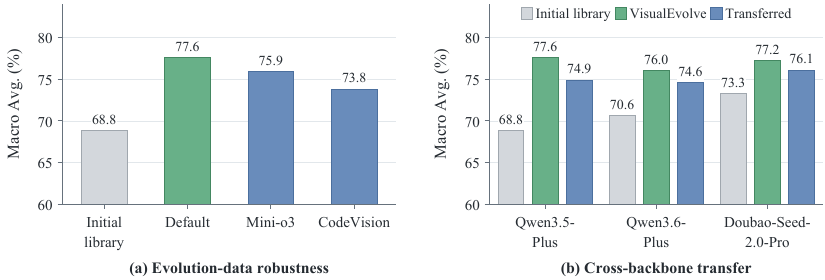}
\caption{\textbf{Generalization of \sys{} (RQ3).}
Macro-average Acc@1 (\%) across six evaluation sets from the three benchmarks.
In (b), \sys{} denotes results from backbone-specific evolution, where the skill library is evolved using the corresponding target backbone.}

\Description{Two side-by-side bar charts report macro-average accuracy.
The left chart shows four individual bars for the Initial Skill Library and the
libraries evolved using Default, Mini-o3, and CodeVision data, all evaluated
with Qwen3.5-Plus.
The right chart compares Qwen3.5-Plus, Qwen3.6-Plus, and Doubao-Seed-2.0-Pro.
Each group shows the Initial Skill Library, VisualEvolve evolved on the target
backbone, and the unchanged library transferred from Kimi-K2.5.
Both y-axes use the same continuous range from 60 to 84 percent;
the absolute accuracy is labeled above every bar.}
\label{fig:generalization-rq3}
\end{figure}

As shown in Figure~\ref{fig:generalization-rq3}, we evaluate VisualEvolve across evolution-data sources and examine cross-backbone transfer of the evolved skill library. For evolution-data sources, we consider two alternatives to the \textit{Default} using Qwen3.5-Plus. \textit{Mini-o3} combines VisualProbe$_{\mathrm{train}}$ (2,900 samples) with DeepEyes-train-4K (4,000 samples)~\citep{minio3}. \textit{CodeVision} combines CodeVision-SFT (6,127 samples) with CodeVision-RL (34,795 samples)~\citep{codevision}.
Only the original
image, question, and answer are retained. For cross-backbone
transfer, we transfer the final skill library evolved with Kimi-K2.5,
$\mathcal{L}_{\mathrm{Kimi}}^{\star}$, to Qwen3.5-Plus, Qwen3.6-Plus, and Doubao-Seed-2.0-Pro.

We make the following observations from Figure~\ref{fig:generalization-rq3}.
(1) VisualEvolve consistently improves macro-average accuracy across all three evolution-data sources, with gains of 8.8, 7.1, and 5.0 percentage points over the Initial Skill Library under Default, Mini-o3, and CodeVision, respectively. These results support the robustness of skill evolution across the data sources.
(2) The transferred skill library $\mathcal{L}_{\mathrm{Kimi}}^{\star}$ improves over the Initial Skill Library on all three other backbones. Its macro-average accuracy is only 2.7, 1.4, and 1.1 percentage points below backbone-specific evolution, respectively. These results show that the evolved skill library transfers effectively across backbones while retaining most of the performance gains from backbone-specific evolution.

\begin{figure}[!t]
\centering
\setlength{\abovecaptionskip}{4pt}
\captionsetup{skip=4pt}
\includegraphics[width=\linewidth]{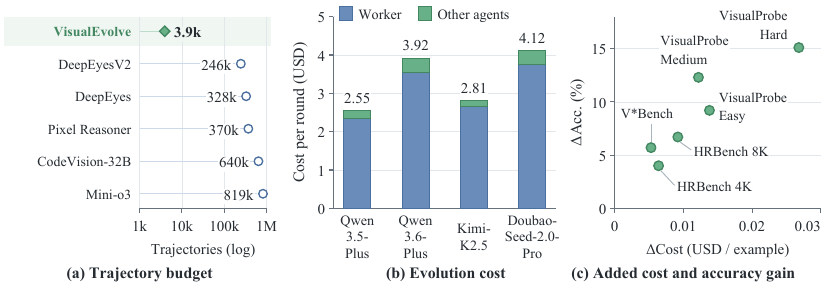}
\caption{\textbf{Evolution budget and inference overhead (RQ4).}
(a) Number of samples processed by the Worker for \sys{} versus estimated RL rollout counts for training-based methods~\citep{deepeyesv2,deepeyes,pixelreasoner,codevision,minio3}, shown on a log scale.
(b) Mean API cost per evolution round.
(c) Added inference cost and absolute accuracy gain over the Initial Skill Library across benchmarks, using Qwen3.5-Plus.}
\Description{Three panels appear in one row.
Panel (a) compares total task executions for VisualEvolve with RL trajectory
counts for five training-based methods, using a logarithmic horizontal axis.
Panel (b) shows stacked per-round API costs for Qwen3.5-Plus, Qwen3.6-Plus,
and Kimi-K2.5, separating Worker and Other agents.
Panel (c) plots the added inference cost in USD per example against the
absolute accuracy gain (\%), with one labeled point for each of
V\textsuperscript{*}Bench, HRBench 4K, HRBench 8K, VisualProbe Easy,
VisualProbe Medium, and VisualProbe Hard.}
\label{fig:cost-adaptation-budget}
\end{figure}

\subsection{Cost Analysis (RQ4)}
\label{app:cost-analysis}

We evaluate the trajectory budget and API cost of evolution, together with the additional inference cost of the evolved library. Details of the trajectory-budget estimation and the cost calculations are provided in Appendix~\ref{app:cost}.

We make the following observations from Figure~\ref{fig:cost-adaptation-budget}.
(1) \sys{} incurs substantially lower cost than training-based methods. It uses only 3,912 Worker task-solving trajectories, whereas the training-based baselines use approximately 63$\times$--209$\times$ more RL trajectories. Moreover, the mean API cost of each evolution round is only \$2.55--\$4.12 across the backbones.
(2) The evolved skill library consistently improves accuracy with only modest additional inference cost. Across all benchmarks, the accuracy gains are accompanied by small increases in cost. Notably, on VisualProbe Hard, an additional cost of only \$0.0268 per example improves accuracy by 15.1 percentage points.
\section{Conclusion}
\label{sec:conclusion}

Self-evolution turns accumulated experience into reusable visual reasoning skills, but multi-round evolution can suffer from performance regressions and evolution evidence overload. VisualEvolve addresses these challenges with Skill-Library Snapshot Selection, which selects the historical snapshot with the highest validation accuracy, and Progressive Evidence Access, which organizes evidence into a coarse-to-fine hierarchy for capability-gap diagnosis. Across four backbones and six evaluation sets from three benchmarks, VisualEvolve improves over the Initial Skill Library in 23 of 24 backbone--evaluation-set combinations, with a maximum gain of 15.1 percentage points on VisualProbe Hard. Ablation studies support the contributions of both Skill-Library Snapshot Selection and Progressive Evidence Access. Further experiments show that VisualEvolve can improve skills using execution experience from different data sources, and that the evolved skill library can be directly reused across VLM backbones.

\subsection*{AI Use Statement}
In this work, we used generative AI tools for prompt refinement and
translation, feedback on research methodology and experimental design,
suggestions on method implementation and result interpretation, and
assistance with parts of our qualitative and thematic data analysis.
We have not used generative AI tools for generating synthetic datasets,
developing theoretical models, formulating mathematical claims, assisting
with mathematical proofs, or cleaning or reformatting datasets.
Additionally, we used generative AI tools for editing parts of the
manuscript to improve readability, summarizing existing literature,
searching for relevant work, and checking references.
We have reviewed all AI-assisted work. Specifically, the authors
evaluated methodological and experimental suggestions, checked
implementation-related suggestions against the actual system and
experimental pipeline, and reviewed AI-assisted analyses, result
interpretations, and manuscript revisions for correctness and consistency
with the reported methods, experimental results, and cited literature. We take responsibility for the final content of this
work, including text, claims or artifacts produced with the aid of
generative AI.

\subsection*{Ethics Statement}

This work does not involve human-subject studies or the collection of new personal or sensitive data. Our experiments evaluate visual reasoning performance using existing
research datasets and vision-language models. We are not
aware of any additional ethical concerns specific to our methodology or experimental setting.

\subsection*{Reproducibility Statement}

The main paper and appendices provide the information needed to
reproduce \sys{}. Section~\ref{sec:method} describes the complete
evolution framework and procedure, while Section~\ref{sec:exp} details
the datasets, sampling protocol, model configurations, baselines,
and benchmarks. Appendix~\ref{app:method-details} provides additional implementation details for the Execute and Evolve stages, including the construction of the Skill-Performance Report and Evolution Log. The appendices also include the initial skill library and the prompts used during skill evolution. 

\FloatBarrier
\bibliographystyle{plainnat}
\bibliography{references}

\iffastcompile
\else
\clearpage
\appendix
\addtocontents{toc}{\protect\setcounter{tocdepth}{1}}
\renewcommand{\contentsname}{Supplementary Material}
\begingroup
\renewcommand{\cfttoctitlefont}{\Large\sffamily\bfseries}
\renewcommand{\cftsecfont}{\normalsize\rmfamily\bfseries}
\renewcommand{\cftsecpagefont}{\normalsize}
\renewcommand{\cftsecleader}{\cftdotfill{\cftdotsep}}
\setlength{\cftbeforesecskip}{5pt}
\tableofcontents
\endgroup
\thispagestyle{fancy}
\vspace{14pt}

\section{Method Details}
\label{app:method-details}

\begingroup
\setlength{\parskip}{3pt}
\setlength{\abovedisplayskip}{5pt plus 1pt minus 2pt}
\setlength{\belowdisplayskip}{5pt plus 1pt minus 2pt}
\setlength{\abovedisplayshortskip}{3pt plus 1pt}
\setlength{\belowdisplayshortskip}{3pt plus 1pt minus 1pt}

This appendix provides the additional
implementation details for Execute and Evolve.

\subsection{Execute Details}
\label{app:execute-details}

This section provides additional implementation details for the Execute stage, including execution set sampling and source references that link trajectory summaries to supporting spans in the original trajectories. 

\noindent\textbf{Balanced Sampling.}
When the evolution data combine multiple datasets, \sys{} applies balanced sampling to ensure that each dataset contributes equally to the evolution evidence.
Before evolution begins, \sys{} constructs the fixed validation set
$\mathcal{D}_{\mathrm{val}}$ using random seed 42 and removes the selected
validation samples from the evolution data.
Then, at each evolution round $r$, Execute uses the round index $r$ as
the random seed and samples 16 examples from each source, yielding a 32-example
execution set $\mathcal{D}_r^{\mathrm{exe}}$.
Therefore, the validation set remains fixed across rounds, while the execution
set is resampled with a round-specific seed, and
$\mathcal{D}_r^{\mathrm{exe}}\cap\mathcal{D}_{\mathrm{val}}=\varnothing$
for every round.

\noindent\textbf{Trajectory Summaries and Source References.}
For a trajectory $T_{r,j}$, the Memory Agent, a trajectory-processing agent, constructs the corresponding trajectory summary $m_{r,j}$. Each trajectory summary contains concise, evidence-supported findings that describe individual behaviors or observations during task execution, including the skills actually loaded,
visual operations invoked, intermediate observations, verification behavior,
failed attempts, and final prediction.
Each finding retains a line-level reference to its supporting span in the
original trajectory. For example, \texttt{[[trajectory.json|L3-L16]]} indicates
that the finding is supported by Lines~3--16 of \texttt{trajectory.json}. The Self-Evolving Agent can follow this reference through a file-reading command that specifies the corresponding start and end lines. Trajectory summaries thereby compress lengthy trajectories while preserving traceable access to the original evidence.
During Evolve, the Self-Evolving Agent reads the summarized task evidence in
$\mathcal{M}_r=\{m_{r,j}\}_{j=1}^{|\mathcal{D}_r^{\mathrm{exe}}|}$ by default. It follows line-level references to the corresponding trajectory segments only when a trajectory summary is insufficient for reliable diagnosis. Complete raw trajectories therefore remain available as fine-grained evidence but are not loaded by default.

\subsection{Evolve Details}
\label{app:evolve-details}

This section provides additional implementation details for the Evolve stage, including post-update structural and executability checks and the construction of the Skill-Performance Report and Evolution Log.

\noindent\textbf{CheckAndRepair.}
After the Self-Evolving Agent updates the skill library, the Orchestrator, a framework-level controller that manages auxiliary procedures, automatically checks its structural validity and executability before the library enters the Evaluate stage. Each skill is stored in a separate directory whose name must match the \texttt{name} field in the YAML frontmatter of \texttt{SKILL.md}. Optional executable visual operations are stored in the \texttt{scripts/} subdirectory and must follow the prescribed interface. Structural checks verify the directory name, YAML frontmatter, Markdown format, and referenced file paths. Executability checks load each optional visual operation through the prescribed interface and perform a basic invocation. If any check fails, Orchestrator returns the failure location and diagnostic message for targeted repair.

\noindent\textbf{Skill-Performance Report Construction.}
After Evaluate assesses a skill-library snapshot $\mathcal{L}_t$ on the fixed
validation set, it records the correctness of each validation example together
with the skills actually loaded. These records form a
validation skill-exposure profile
\[
\mathcal{V}_t =
\left\{
\left(c^{(t)}(d_j), Z_j^{(t)}\right)
\right\}_{d_j\in\mathcal{D}_{\mathrm{val}}},
\]
where $c^{(t)}(d_j)\in\{0,1\}$ indicates whether validation example $d_j$ is
answered correctly under $\mathcal{L}_t$, and $Z_j^{(t)}$ denotes the set of
skills loaded while solving $d_j$.

At the end of round $r$, the Skill-Performance Report $\mathrm{SPR}_r$ is
constructed from the validation skill-exposure profiles of the initial
snapshot $\mathcal{V}_0$, the starting snapshot $\mathcal{V}_{p_r}$, and the
updated snapshot $\mathcal{V}_r$. Following the SPS definition in
Section~\ref{sec:method-evolve}, the report considers the snapshot pairs
$(\mathcal{L}_{p_r},\mathcal{L}_r)$,
$(\mathcal{L}_0,\mathcal{L}_r)$, and
$(\mathcal{L}_0,\mathcal{L}_{p_r})$.
For each pair, SPS is computed only for skills whose corresponding
skill-associated validation subset is non-empty. When snapshot pairs coincide,
the number of distinct comparisons is smaller than three.
The resulting $\mathrm{SPR}_r$ replaces the previous report and is used in round $r+1$.

\noindent\textbf{Evolution Log  Construction.}
The Evolution Log $\mathcal{G}_{r-1}$ records how the skill library changes
across evolution rounds together with the corresponding validation performance.
After evaluating the evolved library $\mathcal{L}_r$, the Orchestrator computes
the difference between the skill-library snapshots from two consecutive rounds,
$\mathcal{L}_{r-1}$ and $\mathcal{L}_r$:
\begin{equation}
\Delta\mathcal{L}_r=
\operatorname{Diff}\!\left(\mathcal{L}_{r-1},\mathcal{L}_r\right).
\label{eq:consecutive-round-diff}
\end{equation}

The resulting per-skill differences are stored as separate \texttt{.diff} files,
each recording the change to a corresponding skill between the two snapshots.
Let $\operatorname{Ref}(\Delta\mathcal{L}_r)$ denote an index of references to
these files. The Orchestrator then constructs an evolution record for round $r$:
\begin{equation}
\rho_r
=
\left(
r,
a_r^{\mathrm{val}},
\operatorname{Ref}(\Delta\mathcal{L}_r)
\right),
\label{eq:evolution-log-record}
\end{equation}
where $r$ is the current round index, $a_r^{\mathrm{val}}$ is the validation accuracy
of $\mathcal{L}_r$, and $\operatorname{Ref}(\Delta\mathcal{L}_r)$ indexes the
corresponding per-skill differences.

The new record $\rho_r$ is appended to the existing log:
\begin{equation}
\mathcal{G}_r
=
\operatorname{Append}(\mathcal{G}_{r-1},\rho_r).
\label{eq:evolution-log-construction}
\end{equation}

Here, $\mathcal{G}_r$ denotes the updated Evolution Log.

\Needspace{6\baselineskip}
\subsection{Complete Evolution Procedure}
\label{app:complete-evolution}
\label{app:detailed-evolution-procedure}

Algorithm~\ref{alg:visualevolve} summarizes the complete evolution procedure
of \sys{}.

\begin{algorithm}[H]
\fontsize{9}{10}\selectfont
\SetAlgoNoLine
\SetAlgoLongEnd
\SetNlSty{textbf}{}{:}
\SetAlgoNlRelativeSize{0}
\SetInd{0.30em}{0.60em}
\DontPrintSemicolon
\caption{\textsc{VisualEvolve}}
\label{alg:visualevolve}

\Require{Frozen backbone agent $\mathcal{A}_{\theta}$; initial skill library
$\mathcal{L}_0$; evolution data $\mathcal{D}_{\mathrm{evo}}$; fixed validation
set $\mathcal{D}_{\mathrm{val}}$; evolution rounds $R$; evaluation sets $\{\mathcal{D}^{eval}_k\}_{k=1}^{K}$}
\Ensure{Final skill library $\mathcal{L}^{\star}$ and benchmark accuracies
$\{\mathrm{Acc}_k^{\mathrm{eva l}}\}_{k=1}^{K}$}

\textbf{Initialization:}\;
$(a_0^{\mathrm{val}},\mathcal{V}_0)\gets
\operatorname{Evaluate}(
\mathcal{A}_{\theta},
\mathcal{L}_0,
\mathcal{D}_{\mathrm{val}})$\;
$\mathcal{C}_0\gets\{\mathcal{L}_0\}$\;
$\mathrm{SPR}_0\gets\varnothing$\;
$\mathcal{G}_0\gets
\operatorname{InitializeEvolutionLog}
(r=0,a_0^{\mathrm{val}})$\;

\For{$r\gets 1$ \KwTo $R$}{

    \textbf{Skill-Library Snapshot Selection:}\;
    Select a starting skill-library snapshot
    $\mathcal{L}_{p_r}\in
    \operatorname*{arg\,max}_{\mathcal{L}_t\in\mathcal{C}_{r-1}}
    a_t^{\mathrm{val}}$\;

    \textbf{Execute:}\;
    $\mathcal{D}_r^{\mathrm{exe}}\gets
    \operatorname{Sample}\!\left(
    \mathcal{D}_{\mathrm{evo}}
    \setminus
    \mathcal{D}_{\mathrm{val}}
    \right)$\;

    $\mathcal{T}_r\gets
    \operatorname{Execute}\!\left(
    \mathcal{A}_{\theta},
    \mathcal{L}_{p_r},
    \mathcal{D}_r^{\mathrm{exe}}
    \right)$\;

    $\mathcal{I}_r\gets
    \operatorname{Compare}\!\left(
    \operatorname{Pred}(\mathcal{T}_r),
    \operatorname{GT}(\mathcal{D}_r^{\mathrm{exe}})
    \right)$\;

    $\mathcal{M}_r\gets
    \operatorname{Summarize}(\mathcal{T}_r)$\;

    \textbf{Evolve:}\;
    $\text{Capability Gap}\gets
    \operatorname{PEA}\!\left(
    \mathcal{L}_{p_r},
    \mathrm{SPR}_{r-1},
    \mathcal{G}_{r-1},
    \mathcal{I}_r,
    \mathcal{M}_r,
    \operatorname{Ref}(\mathcal{T}_r)
    \right)$\;

    $\mathcal{L}_r\gets
    \operatorname{SkillLibraryUpdate}\!\left(
    \mathcal{L}_{p_r},
    \text{Capability Gap}
    \right)$\;

    \textbf{Evaluate:}\;
    $(a_r^{\mathrm{val}},\mathcal{V}_r)\gets
    \operatorname{Evaluate}\!\left(
    \mathcal{A}_{\theta},
    \mathcal{L}_r,
    \mathcal{D}_{\mathrm{val}}
    \right)$\;

    $\mathrm{SPR}_r\gets
    \operatorname{UpdateSkillPerformanceReport}\!\left(
    \mathcal{V}_0,
    \mathcal{V}_{p_r},
    \mathcal{V}_r
    \right)$\;

    $\mathcal{G}_r\gets
    \operatorname{UpdateEvolutionLog}\!\left(
    \mathcal{G}_{r-1},
    \mathcal{L}_{r-1},
    \mathcal{L}_r,
    a_r^{\mathrm{val}}
    \right)$\;

    $\mathcal{C}_r\gets
    \mathcal{C}_{r-1}\cup\{\mathcal{L}_r\}$\;
}

$t^{\star}\in
\operatorname*{arg\,max}_{0\leq t\leq R}
a_t^{\mathrm{val}}$\;

$\mathcal{L}^{\star}\gets
\mathcal{L}_{t^{\star}}$\;

\For{$k\gets 1$ \KwTo $K$}{
    $\mathrm{Acc}_k^{\mathrm{eval}}\gets
    \mathrm{Acc}\!\left(
    \mathcal{A}_{\theta},
    \mathcal{L}^{\star};
    \mathcal{D}_k^{\mathrm{eval}}
    \right)$\;
}

\Return{$\mathcal{L}^{\star}$ and
$\{\mathrm{Acc}_k^{\mathrm{eval}}\}_{k=1}^{K}$}\;

\end{algorithm}
\par\endgroup

\begin{table}[!htbp]
\centering

\fontsize{8}{9.2}\selectfont
\setlength{\tabcolsep}{0.78pt}
\renewcommand{\arraystretch}{1.03}
\begin{tabular}{@{}
  >{\raggedright\arraybackslash}p{0.095\textwidth}
  >{\centering\arraybackslash}p{0.065\textwidth}
  >{\centering\arraybackslash}p{0.065\textwidth}
  >{\centering\arraybackslash}p{0.050\textwidth}
  >{\raggedright\arraybackslash}p{0.165\textwidth}
  >{\raggedright\arraybackslash}p{0.180\textwidth}
  >{\raggedright\arraybackslash}p{0.090\textwidth}
  >{\raggedright\arraybackslash}p{0.262\textwidth}@{}}
\toprule
\textbf{Setting} & \textbf{Round}
& \makecell{\textbf{$\Delta$ Acc.}\\\textbf{(\%)}}
& \textbf{C/I}
& \makecell[l]{\textbf{Diagnosis}}
& \makecell[l]{\textbf{Tool-Use}\\\textbf{Strategy}}
& \makecell[l]{\textbf{Visual}\\\textbf{Operation}}
& \makecell[l]{\textbf{Skill-Loading}\\\textbf{Sequences}} \\
\midrule
\multirow{3}{*}{Default} & r0--r1 & +3.5 & 0/12
& Under-invocation of crop, unstructured observation, and missing answer verification
& $+$\{observe, verify\}; $\sim$\{crop\}
& --
& \textbf{T2:} observe $\rightarrow$ crop (24.5\%) \\
\cmidrule(lr){2-8}
& r1--r2 & +5.5 & 2/12
& Redundant crop operations, incomplete search coverage, and inconsistent counting and spatial reasoning
& $+$\{count, search\}; $\sim$\{crop, observe, verify\}
& --
& \textbf{T2:} observe $\rightarrow$ crop (12.0\%), search $\rightarrow$ crop (7.5\%) \\
\cmidrule(lr){2-8}
& r2--r16 & +4.0 & 5/128
& Persistent errors in color recognition, text reading, counting, localization, and candidate disambiguation
& $+$\{color, read\_text\}; $\sim$\{crop, observe, count, search, verify\}
& --
& \textbf{T2:} observe $\rightarrow$ crop (20.5\%), read\_text $\rightarrow$ crop (13.5\%),
color $\rightarrow$ crop (9.5\%) \\
\midrule
\multirow{2}{*}{CodeVision} & r0--r1 & +4.0 & 12/9
& Rotated or upside-down inputs caused errors in reading text, charts, and numerical values
& $+$\{rotate, measure, verify\}; $\sim$\{crop\}
& rotate$^{\dagger}$
& \textbf{T2:} crop $\rightarrow$ verify (16.5\%), measure $\rightarrow$ crop (11.5\%),
rotate $\rightarrow$ measure (10.5\%) \newline
\textbf{T3:} rotate $\rightarrow$ measure $\rightarrow$ crop (8.0\%),
measure $\rightarrow$ crop $\rightarrow$ verify (8.0\%) \newline
\textbf{T4:} rotate $\rightarrow$ measure $\rightarrow$ crop $\rightarrow$ verify (6.5\%) \\
\cmidrule(lr){2-8}
& r1--r2 & +10.0 & 1/13
& Rotate could not correct mirrored inputs, coordinate and value-reading errors persisted after transformation
& $+$\{flip\}; $\sim$\{crop, measure, rotate, verify\}
& \,\: flip$^{\dagger}$
& \textbf{T2:} rotate $\rightarrow$ flip (18.5\%), crop $\rightarrow$ verify (14.0\%),
flip $\rightarrow$ verify (5.0\%) \newline
\textbf{T3:} rotate $\rightarrow$ flip $\rightarrow$ verify (5.0\%) \\
\bottomrule
\end{tabular}
\par
\caption{\textbf{Skill-evolution dynamics across rounds.} $+$ and $\sim$
denote added and modified tool-use strategies, respectively, while $\dagger$
marks a newly added executable script in the Visual Operation column.
$\Delta$ Acc. is the absolute difference in validation accuracy (\%). C/I denotes the numbers of correct/incorrect Trajectory Summaries inspected by the Self-Evolving Agent, respectively. 
T2, T3, and T4 denote ordered skill-loading sequences containing two, three,
and four skills, respectively; for example, observe $\rightarrow$ crop denotes
a two-skill loading sequence. Parentheses report the fraction of validation
examples in which each sequence occurs, with only rates of at least 5\% shown.}
\label{tab:skill-evolution-behavior}
\end{table}

\section{Performance Gain Analysis}
\label{app:performance-gain-analysis}

\begingroup
\setlength{\intextsep}{5pt}
\setlength{\columnsep}{12pt}
\begin{wrapfigure}{r}{0.50\textwidth}
  \vspace{-5pt}
  \centering
  \includegraphics[width=\linewidth]{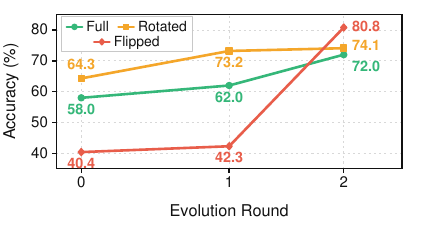}
  \captionsetup{skip=4pt,justification=raggedright}
\caption{\textbf{Validation accuracy on the full CodeVision validation set and its Rotated and Flipped subsets across the first three evolution rounds (\%).} Full denotes the complete CodeVision validation set, while Rotated and Flipped denote the subsets whose original images contain rotation and flipping transformations, respectively.}\label{fig:codevision-transform}
\end{wrapfigure}

To better understand the observed performance gains of \sys{}, we trace the
evolution process from execution trajectories and diagnosed capability gaps,
through skill-library updates, to downstream inference behavior. Table~\ref{tab:skill-evolution-behavior} summarizes representative evolution dynamics under the Default and CodeVision evolution-data settings.

Figure~\ref{fig:codevision-transform} further visualizes the two CodeVision
evolution intervals in the table, showing accuracy changes on the full
validation set and the Rotated and Flipped subsets.

According to Table~\ref{tab:skill-evolution-behavior}
and Figure~\ref{fig:codevision-transform}, we observe that:
(1) Under the default data setting, \sys{} iteratively refines the initial
crop-based strategy into multi-step workflows. The execution trajectories
reveal recurring failures related to insufficient observation, incomplete
search and verification, and ineffective skill invocation. Based on these
capability gaps, \sys{} augments the skill library with skills such as
observe, verify, search, color, and read\_text, while refining the existing
crop skill. These updates are accompanied by improved validation accuracy and
the emergence of multi-step skill-loading sequences, without introducing new
executable visual operations.

(2) Under the CodeVision data setting, \sys{} co-evolves visual operations and tool-use strategies in response to transformation-related failures. Introducing the rotate operation at $r=1$ is accompanied by an increase in accuracy on the Rotated subset from 64.3\% to 73.2\%. After the flip operation is added at $r=2$, accuracy on the Flipped subset further increases from 42.3\% to 80.8\%.
\par\endgroup

\Needspace{220pt}
\section{Execution Set Size Sensitivity}
\label{app:batch-size}

\begingroup
\setlength{\intextsep}{5pt}
\setlength{\columnsep}{12pt}
\begin{wrapfigure}{r}{0.50\textwidth}
  \vspace{-5pt}
  \centering
  \includegraphics[width=\linewidth]{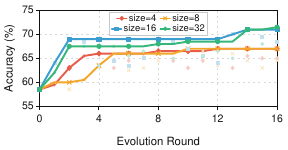}
  \setlength{\abovecaptionskip}{4pt}
  \setlength{\belowcaptionskip}{0pt}
  \caption{\raggedright\textbf{Execution-set size sensitivity.} Lines show the best-so-far
validation accuracy over 16 rounds for execution-set sizes of 4, 8, 16, and
32; scatter points show the corresponding validation accuracy at each round.}
  \Description{A single line chart compares validation accuracy over 16
  evolution rounds when each Execute stage uses 4, 8, 16, or 32 samples.}
  \label{fig:execute-batch-size}
\end{wrapfigure}

We study the sensitivity of \sys{} to the size of $\mathcal{D}_r^{\mathrm{exe}}$
by varying the execution-set size among 4, 8, 16, and 32. All experiments use
Qwen3.5-Plus. Figure~\ref{fig:execute-batch-size} reports the validation
accuracy at each evolution round together with the corresponding best-so-far
accuracy.

We make the following observations from Figure~\ref{fig:execute-batch-size}.
(1) Larger execution sets generally lead to higher best-so-far validation
accuracy. Execution-set sizes of 4 and 8 both reach 67.0\%, while sizes of 16
and 32 reach 71.0\% and 71.5\%, respectively, suggesting that more execution
experience is beneficial for evolution.
(2) The improvement becomes small beyond 16 samples. Increasing the
execution-set size from 16 to 32 raises the best-so-far validation accuracy
only from 71.0\% to 71.5\%. This indicates diminishing returns as the
execution-set size increases.
\par\endgroup

\Needspace{7\baselineskip}
\section{Trajectory-Budget Estimation and API Cost Calculations}
\label{app:cost}

\subsection{Trajectory-Budget Estimation}

We measure the trajectory budget of \sys{} and RL training by the number of task-solving trajectories.

For \sys{}, we count all Worker trajectories generated during skill evolution, including both Execute and validation-time evaluations. Before evolution, the Worker evaluates the initial skill library $\mathcal{L}_0$ on the fixed 200-example validation set. Each of the subsequent $R=16$ evolution rounds generates 32 trajectories during Execute and 200 trajectories during Evaluate. The resulting trajectory budget for one evolution run is
\begin{equation}
\begin{aligned}
N_{\mathrm{\sys}}
&= |\mathcal{D}_{\mathrm{val}}|
 + \sum_{r=1}^{R}\left(|\mathcal{D}_r^{\mathrm{exe}}|
 + |\mathcal{D}_{\mathrm{val}}|\right) \\
&= 200 + (32 + 200)\times 16 = 3{,}912.
\end{aligned}
\label{eq:evolution-trajectory-budget}
\end{equation}
Of these, 512 trajectories come from Execute and 3,400 from validation. This count excludes final benchmark evaluation and non-Worker model calls for trajectory summarization and skill evolution.

For training-based methods, we estimate the number of RL rollout trajectories from the original papers and, when available, official training configurations. For step-based training, we estimate
$N_{\mathrm{RL}}\approx SBG$, where $S$ denotes the number of rollout-collection steps, $B$ the number of input questions per step, and $G$ the number of rollouts generated per question. For epoch-based training, we use
$N_{\mathrm{RL}}\approx EDG$, where $E$ is the number of passes over an RL dataset containing $D$ questions. Table~\ref{tab:trajectory-budget-details} summarizes the estimates used in Figure~\ref{fig:cost-adaptation-budget}(a).

\begin{table}[!htbp]
\centering
\small

\setlength{\tabcolsep}{7pt}
\begin{tabular}{@{}llr@{}}
\toprule
Method & Estimation & RL trajectories \\
\midrule
DeepEyesV2~\citep{deepeyesv2} & $60\times256\times16$ & 245,760 \\
DeepEyes~\citep{deepeyes} & $80\times256\times16$ & 327,680 \\
Pixel Reasoner~\citep{pixelreasoner} & $3\times15{,}402\times8$ & 369,648 \\
CodeVision-32B~\citep{codevision} & $2\times40{,}000\times8$ & $\approx$640,000 \\
Mini-o3~\citep{minio3} & $200\times256\times16$ & 819,200 \\
\bottomrule
\end{tabular}
\par
\caption{\textbf{Estimated trajectory budgets of training-based methods.}
Each estimate follows either $S\times B\times G$ for step-based training or
$E\times D\times G$ for epoch-based training.}
\label{tab:trajectory-budget-details}
\end{table}

For DeepEyesV2, DeepEyes, and Mini-o3, the 60-, 80-, and 200-step rollout horizons are obtained from their reported training curves. Each step uses 256 input questions and 16 rollouts per question; for DeepEyes, these settings are also specified in its official training configuration. For Pixel Reasoner, we use its released 15,402-question RL dataset, three training epochs, and eight rollouts per question from the official training configuration. For CodeVision-32B, we use the approximately 40,000 RL questions reported in the paper, together with two training epochs and eight rollouts per question.

These estimates are intended to approximate task-level trajectory generation rather than total training compute. They exclude SFT, data construction, validation, and additional sampling introduced by filtering. Trajectory length, model size, and parameter-update cost also differ across methods and are therefore not captured by this comparison.

\subsection{API Cost Calculations}

We compute the API cost of \sys{} from the input and output tokens consumed by all model calls during evolution and inference. For each model $m$, let
$p^{m}_{\mathrm{input}}$ and $p^{m}_{\mathrm{output}}$ denote its official non-cache prices per one million input and output tokens at the time of our experiments. The total cost is
\begin{equation}
\mathrm{Cost}_{m}
=
\frac{\mathrm{InputTokens}_{m}}{10^6}p^{m}_{\mathrm{input}}
+
\frac{\mathrm{OutputTokens}_{m}}{10^6}p^{m}_{\mathrm{output}} .
\end{equation}

The input/output prices per one million tokens are 0.8/4.8 RMB for Qwen3.5-Plus, 2.0/12.0 RMB for Qwen3.6-Plus, 4.0/21.0 RMB for Kimi-K2.5, and 3.2/16.0 RMB for Doubao-Seed-2.0-Pro. We do not apply cache-hit discounts. All RMB costs are converted to USD using a fixed exchange rate of 6.82 RMB per USD.

For Figure~\ref{fig:cost-adaptation-budget}(b), we aggregate all API calls within each evolution round and report the mean cost over the 16 rounds. We separately report Worker inference and other evolution calls, including trajectory summarization and skill updates. For Figure~\ref{fig:cost-adaptation-budget}(c), we measure the additional inference cost of the evolved skill library as the difference in mean cost per example between \sys{} and the Initial Skill Library under the same backbone and evaluation set.

\section{Initial Skill Library}
\label{app:initial-skill-library}

Listings~\ref{lst:initial-crop-skill} and~\ref{lst:initial-crop-script} provide the skill instruction and executable visual operation of the initial crop skill, respectively. The former defines the textual strategy $\pi_{\mathrm{crop}}$, while the latter implements the visual operation $u_{\mathrm{crop}}$ for extracting a specified image region. Together, they instantiate $s_{\mathrm{crop}}=(\pi_{\mathrm{crop}},u_{\mathrm{crop}})$, the sole skill in the initial library $\mathcal{L}_0=\{s_{\mathrm{crop}}\}$ introduced in Section~\ref{sec:method-branchable-evolution}. Before the first evolution round, $\mathcal{L}_0$ is evaluated on the fixed validation set and archived as the initial skill library.

\Needspace{7\baselineskip}
\PromptListingTitle{Initial skill definition: \texttt{crop/SKILL.md}.}
\label{lst:initial-crop-skill}
\begin{PromptBlock}
---
name: crop
description: Crop a region for a zoomed sub-view. Use when you need to examine a specific area in greater detail.
---

# Crop

Crop any region of interest using the provided CLI script.

## Command

```bash
python workplace/skills/crop/scripts/crop.py \
  --image <image_path> \
  --bbox <x1>,<y1>,<x2>,<y2> \
  --output <output_path> \
  [--coord-mode relative|absolute]
```

## Arguments

| Argument | Description |
|----------|-------------|
| `--image` | Path to the source image |
| `--bbox` | Bounding box as `x1,y1,x2,y2` (left,top,right,bottom) |
| `--output` | Path to save the cropped image |
| `--coord-mode` | `relative` (default): coordinates in [0,1000] range; `absolute`: pixel coordinates |

## Examples

**Relative coordinates (0-1000):**
```bash
python workplace/skills/crop/scripts/crop.py \
  --image input.jpg \
  --bbox 100,200,500,600 \
  --output artifacts/crop_region1.jpg
```

**Absolute pixel coordinates:**
```bash
python workplace/skills/crop/scripts/crop.py \
  --image input.jpg \
  --bbox 50,100,400,300 \
  --output artifacts/crop_region1.jpg \
  --coord-mode absolute
```

## Output

Prints JSON to stdout:
```json
{"ok": true, "crop_path": "artifacts/crop_region1.jpg", "width": 350, "height": 200}
```

On error:
```json
{"ok": false, "error": "description of the error"}
```

## Tips

- Save cropped images to the `artifacts/` directory for the current sample
- Output picture format is automatically forced to match the input image format (e.g. input `.jpg` -> output `.jpg`).
\end{PromptBlock}

\Needspace{7\baselineskip}
\PromptListingTitle{Initial crop implementation: \texttt{crop/scripts/crop.py}.}
\label{lst:initial-crop-script}
\begin{PromptBlock}[baselinestretch=0.85]
"""CLI crop tool -- Agent calls this via `execute` to crop image regions."""
from __future__ import annotations

import argparse
import json
import sys
from pathlib import Path

from PIL import Image

# The vision model rejects tiny images whose shortest side is <= 10 px,
# so keep the crop content but scale such regions proportionally to 14 px.
MIN_SHORT_SIDE_FOR_MODEL = 14
UPSCALE_TRIGGER_SHORT_SIDE = 10


def _upscale_if_too_small(image: Image.Image) -> Image.Image:
    width, height = image.size
    short_side = min(width, height)
    if short_side > UPSCALE_TRIGGER_SHORT_SIDE or short_side <= 0:
        return image

    scale = MIN_SHORT_SIDE_FOR_MODEL / float(short_side)
    new_width = max(1, int(round(width * scale)))
    new_height = max(1, int(round(height * scale)))
    resampling = getattr(Image, "Resampling", Image).LANCZOS
    return image.resize((new_width, new_height), resample=resampling)


def main() -> None:
    parser = argparse.ArgumentParser(description="Crop a region from an image")
    parser.add_argument("--image", required=True, help="Path to source image")
    parser.add_argument("--bbox", required=True, help="Bounding box: x1,y1,x2,y2")
    parser.add_argument("--output", required=True, help="Path to save cropped image")
    parser.add_argument(
        "--coord-mode",
        default="relative",
        choices=["relative", "absolute"],
        help="Coordinate mode: relative (0-1000) or absolute (pixels)",
    )
    args = parser.parse_args()

    try:
        parts = [s.strip() for s in args.bbox.split(",")]
        if len(parts) != 4:
            raise ValueError(f"bbox must have 4 values, got {len(parts)}")
        x1, y1, x2, y2 = (float(p) for p in parts)

        img = Image.open(args.image)
        w, h = img.size

        if args.coord_mode == "relative":
            # Convert from [0, 1000] relative to pixels
            left = int(x1 * w / 1000)
            top = int(y1 * h / 1000)
            right = int(x2 * w / 1000)
            bottom = int(y2 * h / 1000)
        else:
            left, top, right, bottom = int(x1), int(y1), int(x2), int(y2)

        # Clamp to image bounds
        left = max(0, min(left, w))
        top = max(0, min(top, h))
        right = max(0, min(right, w))
        bottom = max(0, min(bottom, h))

        if left >= right or top >= bottom:
            print(json.dumps({"ok": False, "error": f"Invalid crop box after clamping: ({left},{top},{right},{bottom})"}))
            sys.exit(1)

        cropped = img.crop((left, top, right, bottom))
        cropped = _upscale_if_too_small(cropped)
        out_path = Path(args.output)
        out_path.parent.mkdir(parents=True, exist_ok=True)
        cropped.save(str(out_path))

        cw, ch = cropped.size
        print(json.dumps({"ok": True, "crop_path": str(out_path), "width": cw, "height": ch}))

    except Exception as exc:
        print(json.dumps({"ok": False, "error": str(exc)}))
        sys.exit(1)


if __name__ == "__main__":
    main()
\end{PromptBlock}

\Needspace{7\baselineskip}
\section{Agent Prompts}
\label{app:agent-prompts}

Listings~\ref{lst:worker-prompt}, \ref{lst:memory-agent-prompt},
and~\ref{lst:self-evolving-agent-prompt} provide the complete prompt templates
for the Worker, Memory Agent, and Self-Evolving Agent, respectively.

\Needspace{7\baselineskip}
\subsection{Worker Agent User Prompt}
\label{app:worker-prompt}

Listing~\ref{lst:worker-prompt} presents the user prompt used to guide the Worker during visual task solving. 

\PromptListingTitle{User prompt for the Worker Agent.}
\label{lst:worker-prompt}
\begin{PromptBlock}
# Task
Your task is to answer visual questions about images. Think step by step when solving visual questions.

# Skills
Skills are **external, reusable modules with defined interfaces** that extend your capabilities beyond native reasoning.
Use a skill when the required capability cannot be reliably achieved through direct observation or reasoning:
1. `read_file(<skill_path>/SKILL.md)`
2. Follow the instructions defined in the skill (may include CLI commands)

# Image Inputs and Outputs
- The original image is at `artifacts/original.jpg` and has already been provided in your initial input.
- The folder `artifacts/` already exists. Every image you generate must be saved there and then loaded using `read_file`.

# Constraints
- Use the same file format as the original image for every generated image
  (e.g., original `.jpg` -> outputs must be `.jpg`, not `.png`).
- Do not use tool_call tags such as `<tool_call>...</tool_call>` or `<function=...>`; they are not supported. To use tools, read the relevant skill documentation and follow the applicable instructions.
- Do not access the internet, or any external resource.
- Do not inspect or modify the runtime environment, system configuration, or installed packages and dependencies.
\end{PromptBlock}

\Needspace{7\baselineskip}
\subsection{Memory Agent Task Prompt}
\label{app:memory-agent-prompt}

Listing~\ref{lst:memory-agent-prompt} provides the prompt used by the Memory Agent to compress execution trajectories. It converts each complete trajectory into a set of concise findings covering skill loading, tool use, visual observations, and successful or failed behaviors. Each finding is accompanied by a line-level source reference, allowing the Self-Evolving Agent to access the supporting trajectory segment when the summary alone is insufficient.

\PromptListingTitle{Task prompt for the Memory Agent.}
\label{lst:memory-agent-prompt}
\begin{PromptBlock}
You are **MemoryAgent**. Your role is to maintain structured, evidence-grounded findings about the Worker's visual reasoning trajectory.

## Input
- The full trajectory for a single sample.
- Read it with `read_file("{trajectory_path}")`.

## Trajectory Structure
The trajectory is an ordered list of step records:
- `user`: task description or input
- `assistant`: visible response in `response_content` and, when thinking mode is enabled, optional `reasoning_content` (older trajectories may still use `content`)
- `tool_intent`: planned tool call with arguments
- `tool_result`: returned tool output

## Your Task
Produce a concise structured summary using only explicit evidence from the trajectory.

---

## Requirements
### Findings
Produce evidence-grounded findings.

Each finding must:
- Describe exactly one concrete observation from the trajectory
- Include important tool usage when relevant to the reasoning process
- Include skill loading when the trajectory explicitly shows a skill file being loaded
- Explicitly capture verification, correction, reflection, or failure behaviors when they appear in the trajectory
- End with exactly one citation

Valid citation formats:
- `[[trajectory.json|L<line>]]`
- `[[trajectory.json|L<start>-<end>]]`

---

## Rules
- A skill is considered used only if the trajectory explicitly shows the corresponding skill file being loaded.
- Do not merge multiple observations into a single finding.
- Every finding must be directly supported by its citation.

---

## Output Format
Findings:
- ...
- ...
- ...
\end{PromptBlock}

\Needspace{7\baselineskip}
\subsection{Self-Evolving Agent User Prompt}
\label{app:self-evolving-agent-prompt}

Listing~\ref{lst:self-evolving-agent-prompt} provides the user prompt that guides the Self-Evolving Agent through each evolution round, from Skill-Library Snapshot Selection to Execute, Evolve, and Evaluate. The agent first selects the starting skill-library snapshot and then follows the corresponding Meta-Skills in Appendix~\ref{app:meta-skills}. During Execute and Evaluate, it invokes a script to run the Worker on the execution and validation sets, respectively. During Evolve, it directly diagnoses capability gaps through Progressive Evidence Access and updates the starting skill library accordingly.

\PromptListingTitle{Parameterized user prompt for the Self-Evolving Agent.}
\label{lst:self-evolving-agent-prompt}
\begin{PromptBlock}
# You are a SKILL Optimization Expert.
Your task is to iteratively evolve the Worker's skills for visual reasoning to maximize evaluation accuracy on the fixed validation set.

## 1. File System
Your working directory is the project root. Prefer relative paths
(e.g. `workplace/...`). The `/workplace/...` absolute prefix is also
supported (e.g. `cd /workplace`, `python /workplace/meta-skills/execute/scripts/run_execute.py`).

```
workplace/
|-- skills/                  # Worker skill definitions (SKILL.md)
|-- meta-skills/
|   |-- execute/             # Execute stage
|   |-- evolve/              # Evolve stage
|   `-- evaluate/            # Evaluate stage
|-- memories/                # Current-round per-sample evidence
|   |-- results_index.json   # Correctness index
|   |-- evaluation_metrics_report.md  # Historical evaluation statistics
|   `-- {YYYYMMDD}-{sample_id}/
|       |-- summary.md       # Condensed extraction from trajectory.json with line citations
|       |-- trajectory.json  # Full tool-call trajectory
|       `-- artifacts/       # Saved images from tool calls
|-- evolution_playbook.md    # Code-managed cross-round summary and skill diff index
`-- history/                 # Frozen snapshots per round
    `-- round-N/
        |-- memories/        # Execute-stage evolution evidence snapshot
        |-- context.md       # Full per-round agent context
        |-- skill_diffs/     # Per-skill diffs from round-(N-1) -> round-N
        |-- skills/          # Frozen skills snapshot
        |-- meta-skills/     # Frozen pipeline scripts snapshot
        `-- test_metrics.json  # Evaluate scores
```

## 2. What You Can Evolve
Skills are **external, reusable modules with defined interfaces** that extend worker's capabilities beyond native reasoning. You can evolve files under `workplace/skills/` (modify, add, delete, split, or merge).
- **`skills/{name}/SKILL.md`** -- Worker skill definitions. These are directly loaded and executed by the Worker.
- **`skills/{name}/scripts/`** -- Skill implementation scripts called by the Worker.

## 3. How to Use History
`workplace/history/` stores frozen snapshots for each round. You can:
- Browse scores: `cat workplace/history/round-*/test_metrics.json`
- Inspect historical evidence through
`workplace/memories/evaluation_metrics_report.md` and consecutive-round diffs indexed by `workplace/evolution_playbook.md`.
- Use only current-round `workplace/memories/` for Trajectory Summaries and cited raw-trajectory segments.

## 4. Workflow (per round)
Before each stage, first read the corresponding `meta-skills/<stage_name>/SKILL.md`, then execute the three stages in strict order -- do NOT skip or reorder:
1. **Execute** -- Run workers on training samples.
2. **Evolve** -- Diagnose root causes and improve skills
3. **Evaluate** -- Run the fixed validation set, get validation accuracy

## 5. Constraints
- Do not modify `workplace/meta-skills/` or anything under `workplace/history/`.
- Do not modify `workplace/evolution_playbook.md`; it is fully code-managed by the orchestrator.
- Keep all file operations within `workplace/`. Do not inspect or access anything outside `workplace/`.
- Do not access external resources (including the internet), inspect the runtime environment, or download, install, upgrade, or uninstall any packages or dependencies.

## 6. Round Execution Packet
round_id: {round_id}
[Included when accuracy_history is nonempty]
Accuracy so far:  R{rid_1}={acc_1:.4f}  R{rid_2}={acc_2:.4f} ... 
Round {round_id} is complete only after you finish all three stages in order: execute -> evolve -> evaluate. Your goal is to improve the validation accuracy used for skill-library selection.

[Included only when round_id = 1]
Note: Round 0 was evaluation-only -- no prior execute memories exist. Your Round 1 execute stage produces the first training memories for evolve.

[Included only when round_id > 1]
You can seed this round from a historical skills/ snapshot, using the round with the highest `test_metrics.json` accuracy. After choosing round N, run `rm -rf workplace/skills/* && cp -R workplace/history/round-N/skills/. workplace/skills/`.
\end{PromptBlock}

\Needspace{12\baselineskip}
\section{Meta-Skills}
\label{app:meta-skills}

Listings~\ref{lst:execute-meta-skill}, \ref{lst:evolve-meta-skill},
and~\ref{lst:evaluate-meta-skill} provide the Meta-Skills for Execute,
Evolve, and Evaluate, respectively. Each Meta-Skill specifies the actions
and outputs required at its stage of the workflow outlined by the user
prompt in Listing~\ref{lst:self-evolving-agent-prompt}. The Self-Evolving
Agent autonomously loads and reads the corresponding Meta-Skill before
each stage. These Meta-Skills remain unchanged across evolution rounds. 

\Needspace{7\baselineskip}
\subsection{Execute Meta-Skill}
\label{app:execute-meta-skill}

Listing~\ref{lst:execute-meta-skill} presents the Meta-Skill that guides the Execute stage in constructing the execution evidence described in Section~\ref{sec:method-execute}.

\PromptListingTitle{Meta-Skill for the Execute stage.}
\label{lst:execute-meta-skill}
\begin{PromptBlock}
---
name: execute
description: >-
  USE WHEN: the execute stage of an evolution round.
  NOT FOR: evolve or evaluate stages.
---

# Execute
Collect evidence on how the current skills perform on training samples. The Evolve stage reads this evidence to diagnose failures and improve skills.

## Inputs
- Current skills under `workplace/skills/`.
- Training sample pool (managed by the framework, not directly accessible).

## Workflow
```bash
python workplace/meta-skills/execute/scripts/run_execute.py --round-id <N>
```

- `--round-id` is the only required argument. It must match the current evolution round number so outputs are written to the correct round.
- The script selects the number of samples configured by `execute_num_samples` (default: 32). Samples are randomly drawn from the training pool each round (different from the fixed validation set).

The script runs autonomously; no further action is needed during this stage.

## What the Script Does and Why
For each selected sample:

1. A **Worker agent** attempts the VQA question using skills under `workplace/skills/`. The full execution trace is recorded in trajectory.json.
2. A **Memory agent** is invoked once after the worker finishes, compressing the full trajectory (plus related images) into findings evidence for Evolve: `summary.md` records markdown findings with line citations.

## Outputs
```
workplace/memories/
|-- results_index.json          <- per-sample correctness index
|-- {YYYYMMDD}-sample_001/
|   |-- summary.md              <- Condensed extraction from trajectory.json
|   |-- trajectory.json         <- full tool-call trajectory
|   `-- artifacts/              <- Saved images from tool calls
`-- ...
```

## Constraints
- Do **not** modify any files during this stage -- only run the script.
- After completion, proceed immediately to the **Evolve** stage.
\end{PromptBlock}

\Needspace{7\baselineskip}
\subsection{Evolve Meta-Skill}
\label{app:evolve-meta-skill}

Listing~\ref{lst:evolve-meta-skill} provides the Meta-Skill for diagnosing
capability gaps and updating the skill library through Progressive Evidence
Access (PEA), as described in Section~\ref{sec:method-evolve}.

\PromptListingTitle{Meta-Skill for the Evolve stage.}
\label{lst:evolve-meta-skill}
\begin{PromptBlock}
---
name: evolve
description: >-
  USE WHEN: the evolve stage of an evolution round, after execute stage.
  NOT FOR: execute or evaluate stages.
---

## Evolve
Improve Worker performance on the Validation set by improving Worker skills in `workplace/skills/`, focusing on accuracy.

### Skill Structure
- **YAML frontmatter**:
  - `name`: skill identifier. Use lowercase letters, digits, and hyphens, and keep it identical to the skill directory name.
  - `description`: load trigger. The Worker sees only `name` and `description` before deciding whether to load a skill, so this field must state what the skill does and when to use it.
    - Use valid YAML. Prefer block scalars for descriptions:
      ```yaml
      ---
      name: your-skill-name
      description: >-
        State what the skill does and when the Worker should load it.
      ---
      ```
      If a `description:` value is written on the same line, any unquoted colon inside that value can make the YAML invalid. Use `description: >-` or quote the whole value instead.
- **Body**: detailed guidance visible only after the skill is loaded.
- **Optional scripts**: helper code or utilities used by the skill.

### Inputs
#### Skills to Evolve
- `workplace/skills/{name}/SKILL.md` -- Current skill definition.
- `workplace/skills/{name}/scripts/` -- Current skill scripts (if applicable).

#### Current-Round Memories
- `workplace/memories/results_index.json` -- Ordered sample list (incorrect samples first). Each sample's `memory_dir` field is the path to its `summary.md`.
- `workplace/memories/{YYYYMMDD}-{sample_id}/summary.md` -- Condensed extraction from `trajectory.json` with line citations.
- `workplace/memories/{YYYYMMDD}-{sample_id}/trajectory.json` -- Full execution trace.

#### Skill-Performance Report
- `workplace/memories/evaluation_metrics_report.md` -- Implementation of the Skill-Performance Report, used to identify skills associated with performance changes.

#### Evolution Log
- `workplace/evolution_playbook.md` -- Implementation of the Evolution Log, which records validation accuracy across rounds and indexes the corresponding per-skill diff files.
- `workplace/history/round-N/skill_diffs/{skill}.diff` -- Per-skill diff between the consecutive evaluated snapshots from round `N-1` to round `N`.

### Workflow
#### 1. Analyze Skill-Performance Report
Read `workplace/memories/evaluation_metrics_report.md`.

Use it to identify skills associated with notable performance changes.

#### 2. Analyze Evolution Log
Read `workplace/evolution_playbook.md` to understand validation-accuracy
trends and prior skill changes across rounds.
- Note which update directions were associated with improvements.
- Analyze prior round diffs and validation accuracy to avoid repeating ineffective changes.

#### 3. Analyze Trajectory Summaries
Read `workplace/memories/results_index.json` to see the sample list. Prioritize incorrect samples because they expose skill gaps, but also review correct samples to understand what already works.

Analysis flow:
1. Read the summary file referenced by `memory_dir` first.
   It gives a concise view of the run and includes line-range grounding references.
2. To see the full trajectory details behind the summary, use the line reference in the summary file to read the corresponding range from trajectory.json.
   Example: `read_file("workplace/memories/{YYYYMMDD}-{sample_id}/trajectory.json", offset=41, limit=4)` for `[[trajectory.json|L42-L45]]`.
   If the cited span is too short, expand to nearby lines or continue reading more of `trajectory.json` for context.

For image evidence:
- In step 1, the original image may appear as `{"type": "image_path", "path": "/artifacts/original.jpg"}`.
- For rendered images, look for the `result` field, e.g., `[{"type": "image_path", "path": "artifacts/read_crop_top_right.jpg"}]`.
- You should use `read_file` to inspect these images.

#### 4. Map Failures to Skill Improvements
Map identified issues to targeted changes in `workplace/skills/` (modify, add, delete, split, or merge).

Examples:
- **Load trigger mismatch** -> refine the YAML frontmatter `description` so it clearly states what the skill does and when the Worker should load it.
- **Ambiguity** -> rewrite for clarity and precision
- **Capability gap** -> add new skills (with `mkdir -p`) or supporting logic
- **Redundancy / conflict / bloat** -> delete overlap, resolve contradictions, tighten verbose skills

Skip low-frequency or uncertain issues to avoid overfitting.

### Evolution Heuristics
- **Prioritize generalizable patterns:**
  Because validation samples differ from the training set, avoid overfitting and be cautious with blanket rules like "always" as they rarely generalize.
- **Evidence-driven iteration:**
  Follow `Evidence -> Pattern -> Principle -> Change`.
- **Balance skill length:**
  Excessive length dilutes signal and increases confusion. Overly terse guidance may miss important nuances. Balance requires cross-round analysis.
- **Avoid overloading a single skill:**
  An overloaded skill blurs trigger boundaries and leads to irrelevant or conflicting guidance. Instead, add dedicated skills for each question type or scenario, with clear trigger conditions.

### Outputs
- Updated files in `workplace/skills/`:
  - `SKILL.md`
  - related scripts, if applicable

### Constraints
- Changes are allowed only in `workplace/skills/`.
- Use:
  - `edit_file` for targeted edits.
  - `write_file` for full rewrites.
\end{PromptBlock}

\Needspace{12\baselineskip}
\subsection{Evaluate Meta-Skill}
\label{app:evaluate-meta-skill}

Listing~\ref{lst:evaluate-meta-skill} provides the Meta-Skill for assessing
the updated skill library during Evolve, following the procedure in
Section~\ref{sec:method-evaluate}.

\PromptListingTitle{Meta-Skill for the Evaluate stage.}
\label{lst:evaluate-meta-skill}
\begin{PromptBlock}
---
name: evaluate
description: >-
  USE WHEN: the evaluate stage of an evolution round, after evolve stage.
  NOT FOR: execute or evolve stages.
---

# Evaluate
Measure whether the skill changes from Evolve actually generalize. Accuracy on the fixed validation set is the definitive metric -- it answers "did we improve, regress, or stay flat?"

## Inputs
- Worker skills under `workplace/skills/`.
- Fixed validation set (managed by the orchestrator).

## Workflow
```bash
python workplace/meta-skills/evaluate/scripts/run_evaluate.py \
  --round-id <N>
```

The script runs the worker on every validation sample and scores each answer via judge.

After the script finishes, workplace/history/round-{N}/test_metrics.json is written.

## Outputs
- `workplace/history/round-{N}/test_metrics.json` -- Validation accuracy for this round.

## Constraints
- Do **not** change anything under `workplace/`.
- Run `run_evaluate.py` at most once per round.
- After completion, this round is done. The orchestrator will start the next round.
\end{PromptBlock}

\clearpage
\section{Case Studies}
\label{app:case-studies}

We provide two case studies to illustrate how the evolved skill library supports visual reasoning on individual tasks. Figure~\ref{fig:case-operation-composition} shows a reasoning trajectory in which the evolved library composes the rotate and flip operations to handle an image requiring multiple transformations. Figure~\ref{fig:case-iterative-inspection} shows a reasoning trajectory in which the evolved skill library combines skill guidance with progressively tighter crops and verification to gather sufficient visual evidence.

\noindent\begin{minipage}{\linewidth}
  \centering
  \includegraphics[width=0.95\textwidth,height=0.77\textheight,keepaspectratio]
  {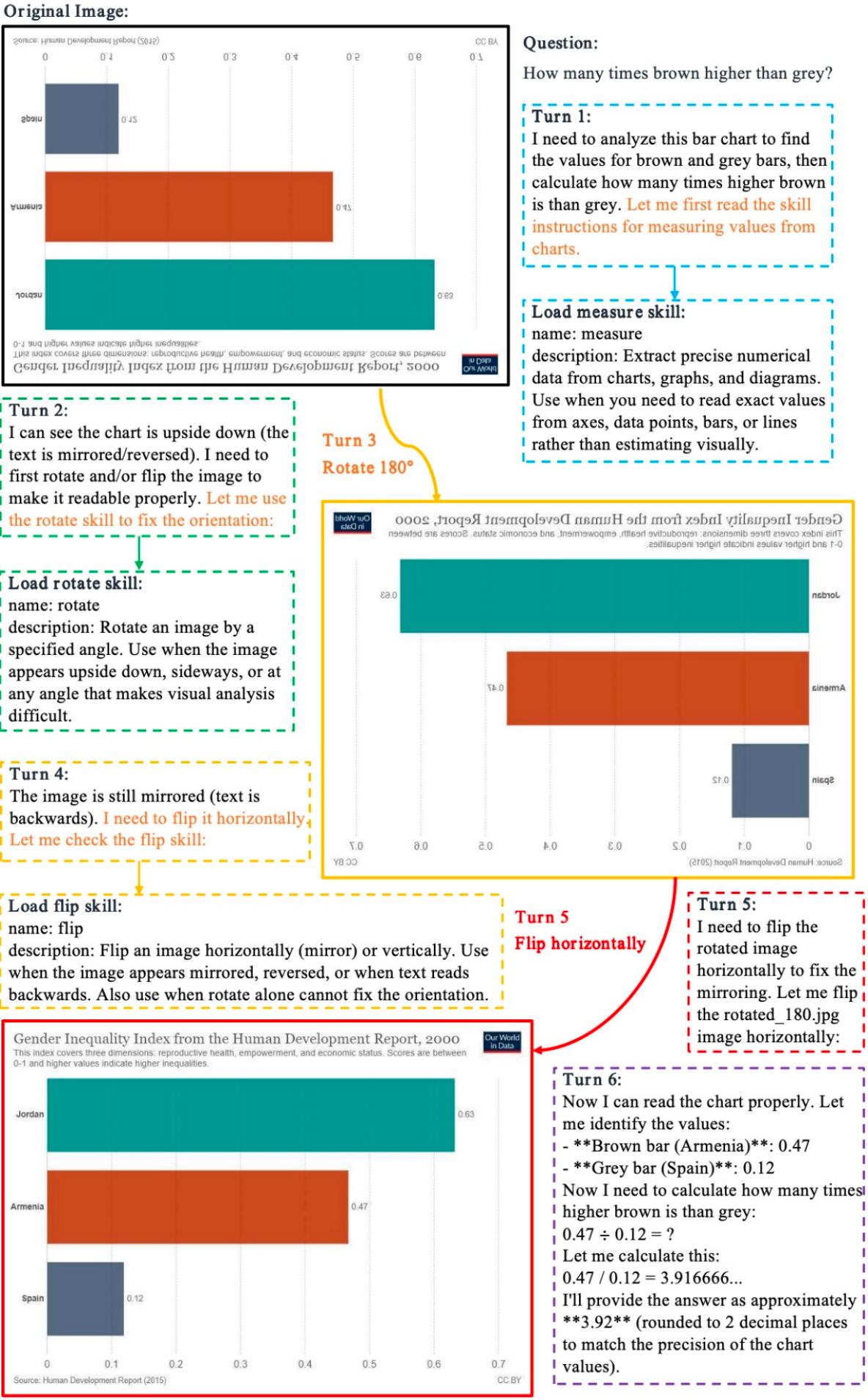}
  \captionof{figure}{\textbf{Case study of visual-operation expansion and composition.}
  The CodeVision-evolved library first loads the measure skill, then recognizes
  that the chart is both upside down and mirrored. It composes the newly
  discovered rotate and flip operations to recover a readable chart before
  extracting the values and computing the final answer.}
  \Description{A visual reasoning trajectory for an inverted and mirrored bar
  chart. The agent loads measure, rotate, and flip skills, transforms the
  image, reads the corrected chart, and calculates the answer.}
  \label{fig:case-operation-composition}
\end{minipage}

\clearpage
\noindent\begin{minipage}{\linewidth}
  \centering
  \includegraphics[width=0.95\textwidth,height=0.91\textheight,keepaspectratio]
  {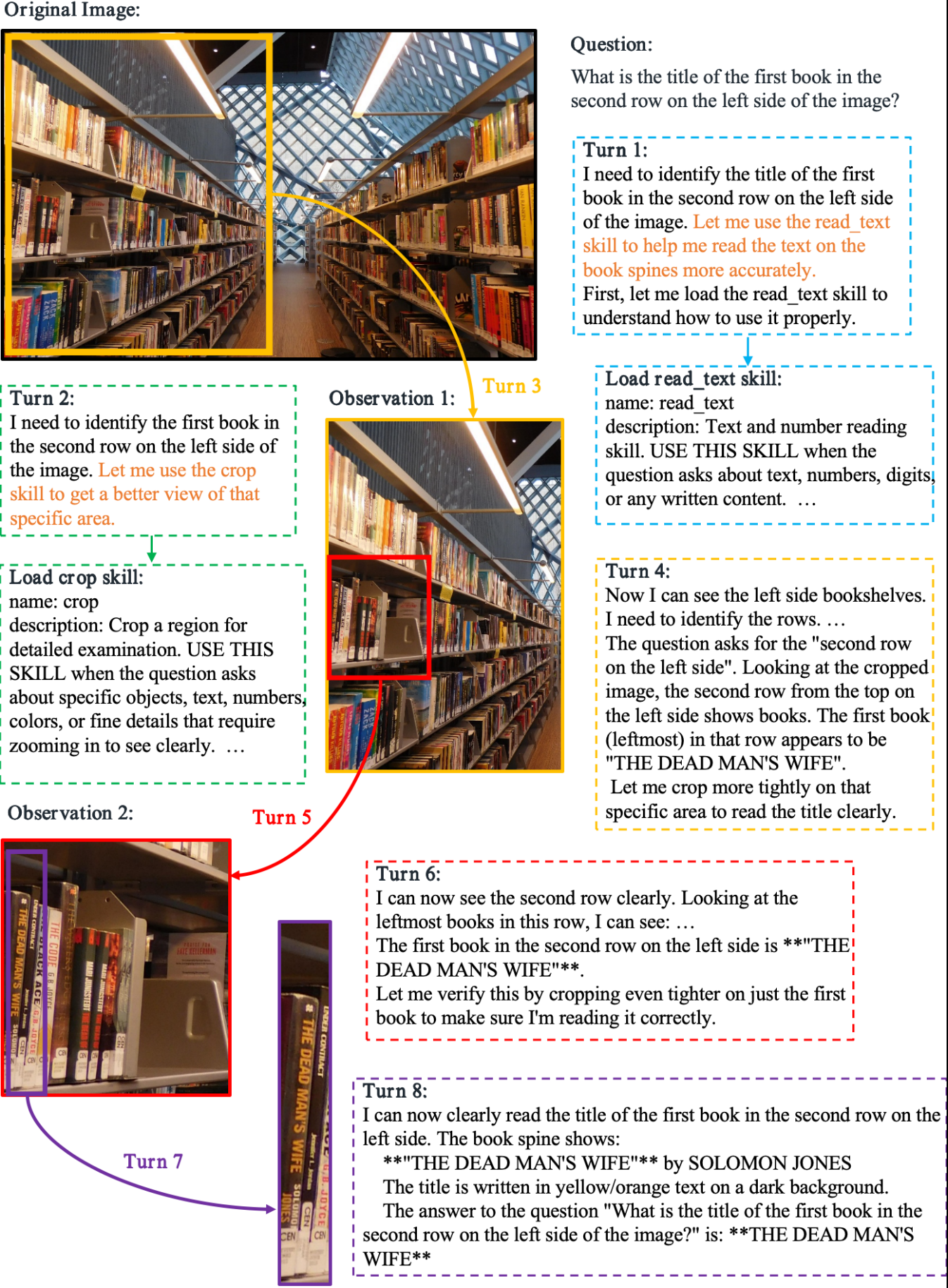}
  \captionof{figure}{\textbf{Case study of iterative local inspection and verification.}
  The evolved library combines read\_text guidance with progressively tighter
  crops. The agent first localizes the relevant bookshelf region, then
  repeatedly narrows the visual evidence and verifies the target book spine
  before returning the title.}
  \Description{A visual reasoning trajectory for reading a book title in a
  library image. The agent loads read-text and crop skills, performs
  progressively tighter crops, verifies the book spine, and answers with the
  title.}
  \label{fig:case-iterative-inspection}
\end{minipage}

\clearpage

\fi

\end{document}